%% file: paper.tex
\documentclass[10pt,conference]{IEEEtran}

\usepackage{booktabs} 
\usepackage{multirow}

\input{preamble}

\begin{document}

\title{\LARGE \emph{To Compress, or Not to Compress:} Characterizing Deep Learning Model Compression for Embedded Inference \vspace{-4mm}}

\author{\IEEEauthorblockN{Qing Qin\IEEEauthorrefmark{3},
         Jie Ren\IEEEauthorrefmark{2}, Jialong Yu\IEEEauthorrefmark{3}, Ling Gao\IEEEauthorrefmark{3}, Hai Wang\IEEEauthorrefmark{3}, Jie
         Zheng\IEEEauthorrefmark{3},
         Yansong Feng\IEEEauthorrefmark{4},
         Jianbin Fang\IEEEauthorrefmark{5},
         Zheng Wang\IEEEauthorrefmark{1}}

 \IEEEauthorblockA{\IEEEauthorrefmark{3} Northwest University, China, \IEEEauthorrefmark{2} Shaanxi Normal University, China, \IEEEauthorrefmark{4} Peking University, China}
 \IEEEauthorblockA{\IEEEauthorrefmark{5} National University of Defense Technology, China, \IEEEauthorrefmark{1} Lancaster University, United Kingdom}
\vspace{-15mm}
 }

\maketitle

\input{abstract}


\input{intro}

\input{background}
\input{setup}
\input{results}
\input{discussions}
\input{related_work}
\input{conclusions}
\input{ack}

\bibliographystyle{plain}
\bibliography{refs}
\balance

\end{document}

%% file: preamble.tex
\usepackage{graphicx}
\usepackage{caption}

\usepackage{subfig}
\pdfoutput=1

\usepackage{blindtext}
\usepackage{adjustbox}
\usepackage{multirow}
\usepackage{color}
\usepackage{booktabs}
\usepackage{tabularx}
\usepackage{colortbl}
\usepackage{bbding}
\usepackage{tikz}
\usepackage{listings}
\usepackage{etoolbox}
\usepackage{subfig}

\usepackage{url}
\usepackage{setspace}
\usepackage[american,english]{babel}
\usepackage{wrapfig}
\usepackage{tabularx}
\usepackage{balance}

\newcommand{\circled}[2][]{\tikz[baseline=(char.base)]
    {\node[shape = circle, draw, inner sep = 1pt]
    (char) {\phantom{\ifblank{#1}{#2}{#1}}};%
    \node at (char.center) {\makebox[0pt][c]{#2}};}}
\robustify{\circled}

\makeatletter
\def\@IEEEsectpunct{.\ \,}
\def\paragraph{\@startsection{paragraph}{4}{\z@}{1.5ex plus 1.5ex minus 0.5ex}%
{0ex}{\normalfont\normalsize\sffamily\bfseries}}

\patchcmd{\@maketitle}
  {\addvspace{0.2\baselineskip}\egroup}
  {\addvspace{-1\baselineskip}\egroup}

\makeatother

\newcommand{\cparagraph}[1]{\vspace{1.5mm}\noindent \textbf{#1}}

\newcommand{\DNN} {\texttt{DNN}\xspace}
\newcommand{\DNNs} {\texttt{DNNs}\xspace}
\newcommand{\CNN} {\texttt{CNN}\xspace}
\newcommand{\RNN} {\texttt{RNN}\xspace}
\newcommand{\RNNs} {\texttt{RNNs}\xspace}
\newcommand{\CNNs} {\texttt{CNNs}\xspace}

\newcommand{\pruning} {\emph{pruning}\xspace}
\newcommand{\quantization} {\emph{quantization}\xspace}
\newcommand{\dquantization} {\emph{data quantization}\xspace}

\newcommand*{\eg}{\textit{e.g.}\@\xspace}

\newcommand*{\etal}{%
    \@ifnextchar{.}%
        {et al}%
        {et al.\@\xspace}%
}

\definecolor{Gray}{gray}{0.9}

%% file: abstract.tex
\begin{abstract}
The recent advances in deep neural networks (\DNNs) make them attractive for embedded systems. However, it can take a long time for DNNs
to make an inference on resource-constrained computing devices. Model compression techniques can address the computation issue of deep
inference on embedded devices. This technique is highly attractive, as it does not rely on specialized hardware, or
computation-offloading that is often infeasible due to privacy concerns or high latency. However, it remains unclear how model
compression techniques perform across a wide range of \DNNs. To design efficient embedded deep learning solutions, we need to understand
their behaviors. This work develops a quantitative approach to characterize model compression techniques on a representative embedded
deep learning architecture, the NVIDIA Jetson Tx2. We perform extensive experiments by considering 11 influential neural network
architectures from the image classification and the natural language processing domains. We experimentally show that how two mainstream
compression techniques, data quantization and pruning, perform on these network architectures and the implications of compression
techniques to the model storage size, inference time, energy consumption and performance metrics. We demonstrate that there are
opportunities to achieve fast deep inference on embedded systems, but one must carefully choose the compression settings. Our results
provide insights on when and how to apply model compression techniques and guidelines for designing efficient embedded deep learning
systems.
\end{abstract}

%% file: intro.tex
\section{Introduction}
In recent years, deep learning has emerged as a powerful tool for solving complex problems that were considered to be difficult in the
past. It has brought a step change in the machine's ability to perform tasks like object recognition~\cite{donahue14,he2016deep}, facial
recognition~\cite{parkhi2015deep,sun2014deep}, speech processing~\cite{pmlrv48amodei16}, and machine translation~\cite{bahdanau2014neural}.
While many of these tasks are also important on mobiles and the Internet of Things (IoT), existing solutions are often
computation-intensive and require a large amount of resources for the model to operate. As a result, performing deep
inference\footnote{Inference in this paper refers to apply a pre-trained model on an input to obtain the corresponding output. This is
different from statistical inference.} on embedded devices can lead to long runtime and the consumption of abundant amounts of resources,
including CPU, memory, and power, even for simple tasks~\cite{CanzianiPC16}. Without a solution,
 the hoped-for advances on embedded sensing will not arrive.

Numerous approaches have been proposed to accelerate deep inference on embedded devices. These include designing purpose-built hardware to
reduce the computation or memory latency~\cite{georgiev2017low}, compressing a pre-trained model to reduce its storage and memory footprint
as well as computational requirements~\cite{Han:2016:EEI:3001136.3001163}, and offloading some, or all, computation to a cloud
server~\cite{Kang2017neurosurgeon,teerapittayanon2017distributed}. Compared to specialized hardware, model compression techniques have the
advantage of being readily deployable on commercial-off-the-self hardware; and compared to computation offloading, compression enables
local, on-device inference which in turn reduces the response latency and has fewer privacy concerns. Such advantages make model
compressions attractive on resource-constrained embedded devices where computation offloading is infeasible.

However, model compression is not a free lunch as it comes at the cost of loss in prediction accuracy~\cite{Cheng2017A}. This means that
one must carefully choose the model compression technique and its parameters to effectively trade precision for time, energy, as well as
computation and resource requirements. Furthermore, as we will show in this paper, the reduction in the model size does not necessarily
translate into faster inference time. Because there is no guarantee for a compression technique to be profitable, we need to know when and
how to apply a compression technique.

Our work aims to characterize deep learning model compression techniques for embedded inference. Knowing this not only assists the better
deployment of computation-intensive models, but also informs good design choices for deep learning models and accelerators.

To that end, we develop a quantitative approach  to characterize two mainstream model compression techniques, pruning~\cite{Cheng2017A} and
data quantization~\cite{Gong2014Compressing}. We apply the techniques to the image classification and the natural language processing (NLP)
domains, two areas where deep learning has made great breakthroughs and a rich set of pre-trained models are available. We evaluate the
compression results on the NVIDIA Jetson TX2 embedded deep learning platform and consider a wide range of influential deep learning models
including convolutional and recurrent neural networks. 

We show that while there is significant gain for choosing the right compression technique and parameters, mistakes can seriously hurt the
performance. We then quantify how different model compression techniques and parameters affect the inference time, energy consumption,
model storage requirement and prediction accuracy. As a result, our work provides insights on when and how to apply deep learning model
compression techniques on embedded devices, as well as guidelines on designing schemes to adapt deep learning model optimisations for
various application constraints.

\begin{figure*}[!t]
\centering
\subfloat[][Model size]{\includegraphics[width=0.22\textwidth]{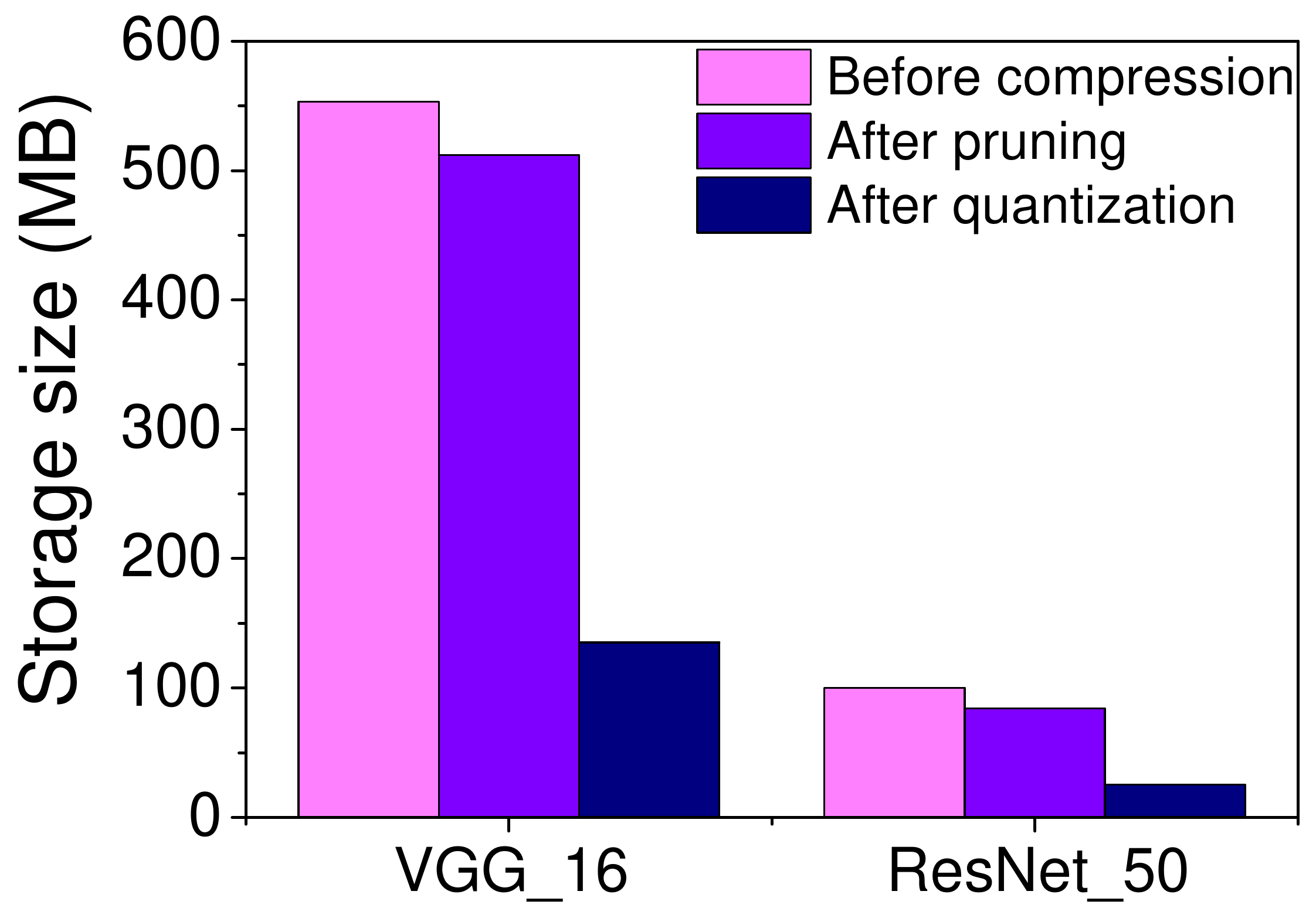}}
\hfill
\subfloat[][Inference time]{\includegraphics[width=0.22\textwidth]{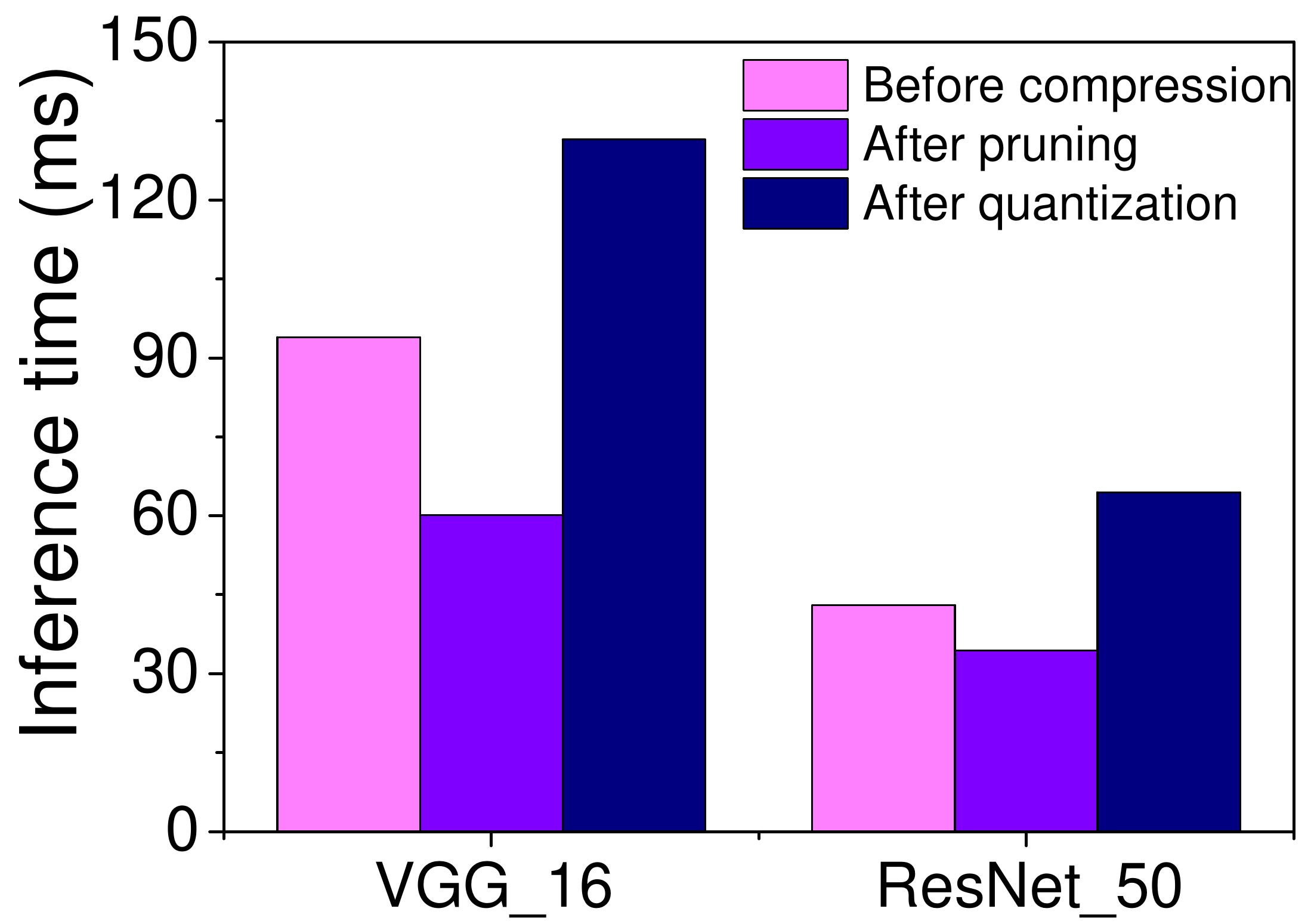}}
\hfill
\subfloat[][Energy consunption]{\includegraphics[width=0.21\textwidth]{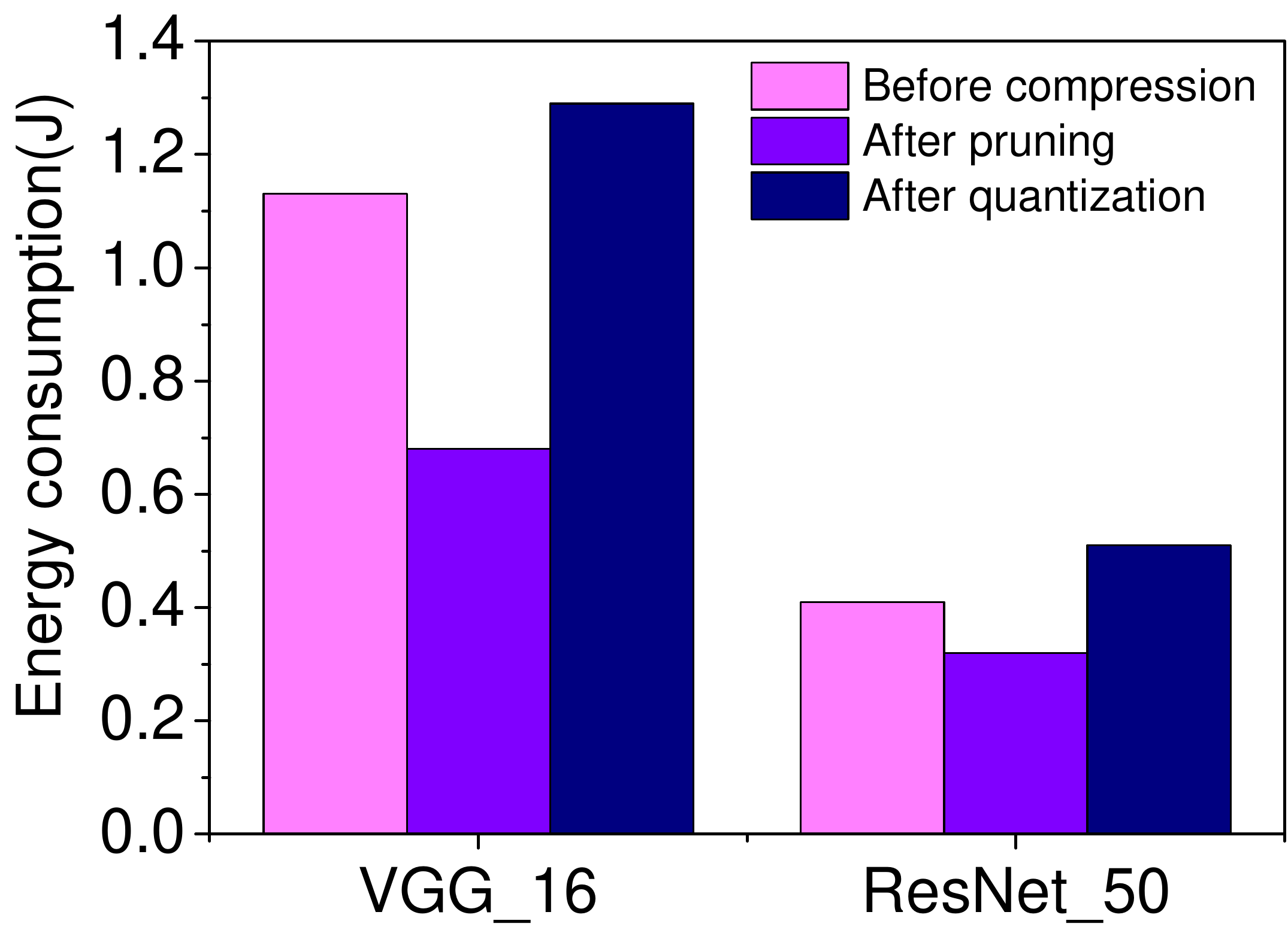}}
\hfill
\subfloat[][Accuracy]{\includegraphics[width=0.212\textwidth]{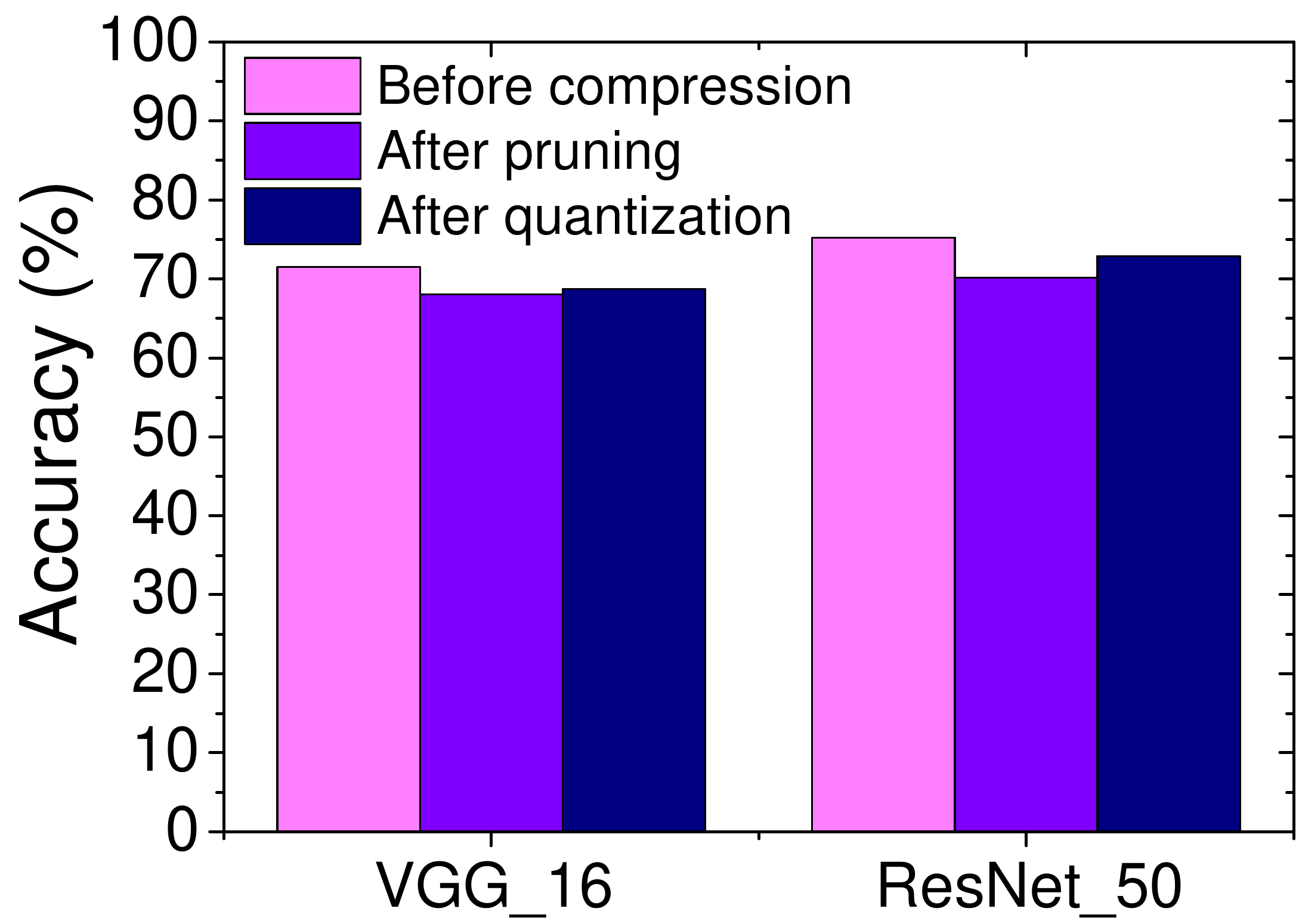}}
\hfill
\caption{The achieved model size (a) inference time (b) energy consumption (c) and accuracy (d) before and after the compression by \quantization and \pruning.
The compression technique to use depends on the optimization target.}
\vspace{-6mm}
\label{fig:motivation}
\end{figure*}

The main contributions of this \emph{workload characterization} paper are two folds:

\begin{itemize}
\item We present the first comprehensive study for deep learning model compression techniques on embedded systems;
\item Our work offers new insights on when and how to apply compression techniques for embedded deep inference.
\end{itemize}


%% file: background.tex
\section{Background and Motivation}
\subsection{Background}
In this work, we consider two commonly used model compression techniques, described as follows.

\cparagraph{Pruning.} This technique removes less important parameters and pathways from a trained network. Pruning ranks the neurons in
the network according how much the neuron contribute, it then removes the low ranking neurons to reduce the model size. Care must be taken
to not remove too many neurons to significantly damage the accuracy of the network.

\cparagraph{Data quantization.} This technique reduces the number of bits used to store the weights of a network, e.g., using 8 bits to
represent a 32-bit floating point number. In this work, we apply data quantization to convert a pre-trained floating point model into a
fixed point model without re-training. We use 6, 8 and 16 bits to represent a 32-bit number, as these are the most common fixed-pointed
data quantization configurations~\cite{pacq}.

\subsection{Motivation}
Choosing the right compression technique is non-trivial. As a motivation example, consider applying \pruning and \dquantization, to two
representative convolutional neural networks (\CNN), \texttt{VGG\_16} 	and \texttt{Resnet\_50}, for image classification. Our evaluation
platform is a NVIDIA Jetson TX2 embedded platform (see Section~\ref{sec:platform}).

\cparagraph{Setup.} We apply each of the compression techniques to the pre-trained model (which has been trained on the ImageNet ILSVRC
2012 training dataset~\cite{imagenet2012}). We then test the original and the compressed models on ILVRSC 2012 validation set which
contains 50k images. We use the GPU for inferencing.

\cparagraph{Motivation Results.} Figure~\ref{fig:motivation} compares the model size, inference time, energy consumption and accuracy after
applying compression. By removing some of the nerons of the network, \pruning is able to reduces the inference time and energy consumption
by 28\% and 22.5\%, respectively. However, it offers little saving in storage size because network weights still dominate the model size.
By contrast, by using a few number of bits to represent the weights, \quantization significantly reduces the model storage size by 75\%.
However, the reduction in the model size does not translate to faster inference time and fewer energy consumption; on the contrary, the
inference time and energy increase by 1.41x and 1.19x respectively. This is because the sparsity in network weights brought by
\quantization leads to irregular computation which causes poor GPU performance~\cite{DBLP:journals/corr/abs-1802-10280} and the cost of
de-quantization (details in Section~\ref{sec:time}) . Applying both compression techniques has modest impact on the prediction accuracy, on
average, less than 5\%. This suggests that both techniques can be profitable.

\cparagraph{Lessons Learned.} This example shows that the compression technique to use depends on what to be optimized for. If storage
space is a limiting factor, \quantization provides more gains over \pruning, but a more powerful processor unit is required to achieve
quick on-device inference. If faster on-device turnaround time is a priority, \pruning can be employed but it would require sufficient
memory resources to store the model parameters.  As a result, the profitability of the compression technique depends on the optimization
constraints. This work provides an extensive study to characterize the benefits and cost of the two model compression techniques.

%% file: setup.tex
\section{Experimental Setup \label{sec:setup}}
\subsection{Platform and Models\label{sec:platform}}

\begin{table}[t!]
\begin{center}
\vspace{-1mm}
\caption{List of deep learning models considered in this work.}
\vspace{-2mm}
\scriptsize
\label{tab:workload}
\begin{tabularx}{\columnwidth}{llXXrr}
\toprule
\textbf{Model}&\textbf{Type}& \textbf{Top-1 (\%)} & \textbf{Top-5 (\%)}& \textbf{\#param.s}& \textbf{Depth} \\
\midrule
\rowcolor{Gray} \textbf{NMT} & \texttt{RNN}  & 27.4 (BLEU)	& -	&211M	&4 \\
\textbf{Inception\_v1}     &\CNN & 69.8  & 89.6	&7M&	22  \\
\rowcolor{Gray} \textbf{Inception\_v2}     &\CNN &73.9	&91.4 & 11.3M& 	32 \\
\textbf{Inception\_v3}     &\CNN  & 78	&94 &	27.1M	&42 \\
\rowcolor{Gray} \textbf{Inception\_v4}     &\CNN& 80.2	&95.2 & 25.6M&58 \\
\textbf{ResNet\_50}        &\CNN & 75.2&	90.2&	25.5M &	50\\
\rowcolor{Gray} \textbf{ResNet\_101} &\CNN &76.4	&92.9&	51M&	101  \\
\textbf{ResNet\_152} &\CNN & 76.8	&93.2&	76.5M	&152\\
\rowcolor{Gray} \textbf{VGG\_16} &\CNN, fully conn. & 71.5&	89.8	&138M	&16 \\
\textbf{VGG\_19} &\CNN, fully conn. & 71.1	&89.8	&138M	&19  \\
\rowcolor{Gray} \textbf{MobileNet} & \CNN & 70.7 & 89.56 &	4.2M	&28 \\

\bottomrule
\end{tabularx}
\end{center}
\vspace{-8mm}
\end{table}

\cparagraph{Hardware.} Our experimental platform is the NVIDIA Jetson TX2 embedded platform. The system has a 64~bit dual-core Denver2 and
a 64~bit quad-core ARM Cortex-A57 running at 2.0~Ghz, and a 256-core NVIDIA Pascal GPU running at 1.3~Ghz. The board has 8~GB of LPDDR4 RAM
and 96~GB of storage (32~GB eMMC plus 64~GB SD card).

\cparagraph{System Software.} We run the Ubuntu 16.04 operating system with Linux kernel v4.4.15. We use Tensorflow v.1.6, cuDNN (v6.0) and
CUDA (v8.0.64).

\cparagraph{Deep Learning Models.} We consider 10 pre-trained \CNN models for image recognition from the TensorFlow-Slim
library~\cite{silberman2013tensorflow} and a recurrent neural network (\RNN) model for machine translation. Table~\ref{tab:workload} lists
the models considered in this work. The chosen models have different parameter sizes and network depths, and thus cover a wide range of
\CNN and \RNN model architectures. We apply \dquantization to \CNN models because the current Tensorflow implementation does not support
quantization of \RNNs. As \pruning requires model updates through retraining, we consider three typical models for \pruning to keep the
experiment manageable.

\subsection{Evaluation Methodology \label{sec:method}}

\cparagraph{Performance Metrics} We consider the following metrics:
\begin{itemize}
\item \emph{\textbf{Inference time} (lower is better)}. Wall clock time between a model taking in an input and producing an output,
    excluding the model load time.

\item \emph{\textbf{Power/Energy consumption} (lower is better)}. The energy used by a model for inference.  We deduct the static power used by
    the hardware when the system is idle.

\item \emph{\textbf{Accuracy} (higher is better)}. The ratio of correctly labeled images to the total number of testing instances.

\item \emph{\textbf{Precision} (higher is better)}. The ratio of a correctly predicted instances to the total number of instances that
    are predicted to have a specific label. This metric answers e.g., ``\emph{Of all the images that are labeled to have a cat, how many
    actually have a cat?}".

\item \emph{\textbf{Recall} (higher is better)}. The ratio of correctly predicted instances to the total number of test instances that
    belong to an object class. This metric answers e.g., ``\emph{Of all the test images that have a cat, how many are actually labeled to
    have a cat?}".

\item \emph{\textbf{F1 score} (higher is better)}.  The weighted average of Precision and Recall, calculated as $2\times\frac{Recall
    \times Precision} {Recall + Precision}$. It is useful when the test dataset has an uneven distribution of classes.

\item \emph{\textbf{BLEU} (higher is better)}. The bilingual evaluation understudy (BLEU) evaluates the quality of machine translation.
    The quality is considered to be the correspondence between a machine's output and that of a human: ``\emph{the closer a machine
    translation is to a professional human translation, the better it is}". We report the BLUE value on \texttt{NMT}, a machine
    translation model.

\end{itemize}

\cparagraph{Performance Report.}   For image recognition, the accuracy of a model is evaluated using the top-1 score by default; and we
also consider the top-5 score. We use the definitions given by the ImageNet Challenge. Specifically, for the top-1 score, we check if the
top output label matches the ground truth label of the primary object; and for the top-5 score, we check if the ground truth label of the
primary object is in the top 5 of the output labels for each given model. For NLP, we use the aforementioned BLEU metric. Furthermore, to
collect inference time and energy consumption, we run each model on each input repeatedly until the 95\% confidence bound per model per
input is smaller than 5\%. In the experiments, we exclude the loading time of the \CNN models as they only need to be loaded once in
practice. To measure energy consumption, we developed a runtime to take readings from the on-board power sensors at a
frequency of 1,000 samples per second. We matched the power readings against the time stamps of model execution to calculate the
energy consumption, while the power consumption is the mean of all the power readings.

%% file: results.tex
\begin{figure*}
\centering
\begin{minipage}[c]{0.7\textwidth}
\centering
\includegraphics[width=1\textwidth]{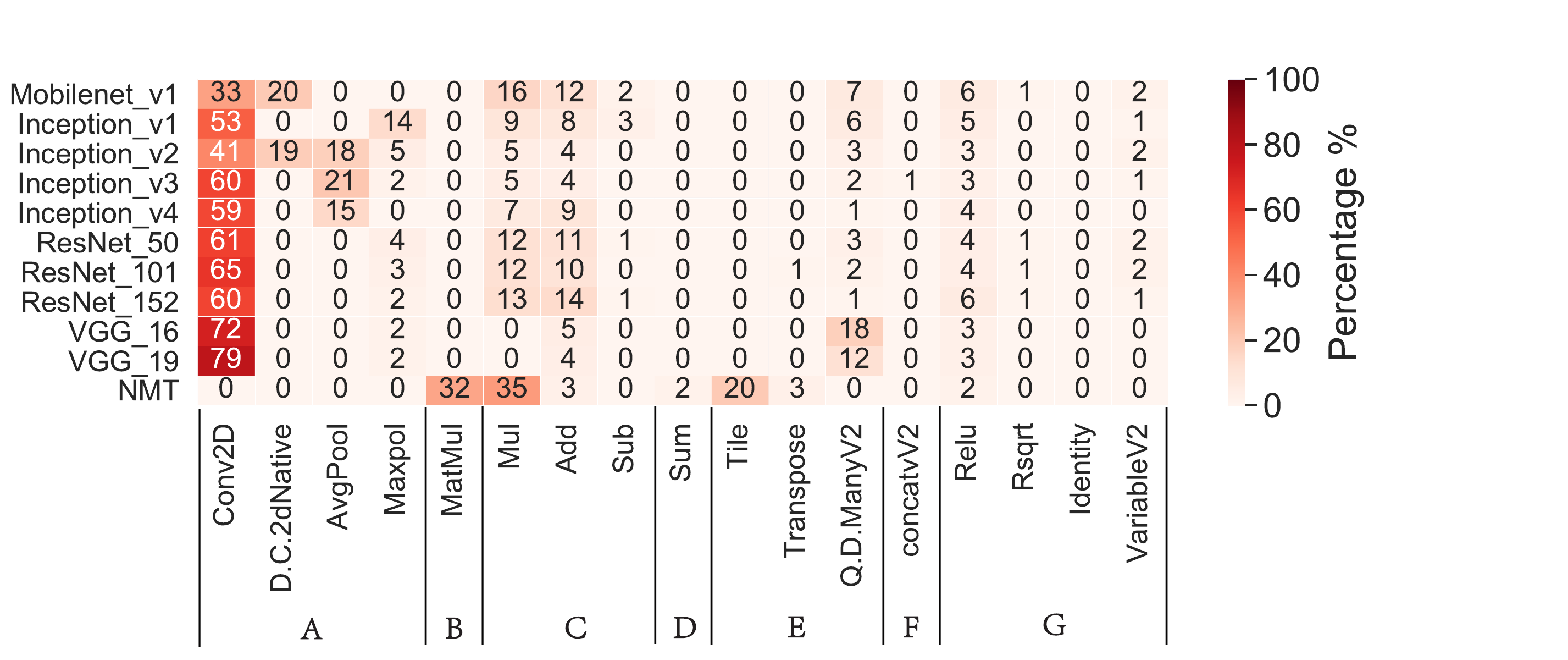}
\end{minipage}%
\begin{minipage}[c]{0.13\textwidth}
\centering
\includegraphics[width=0.9\textwidth]{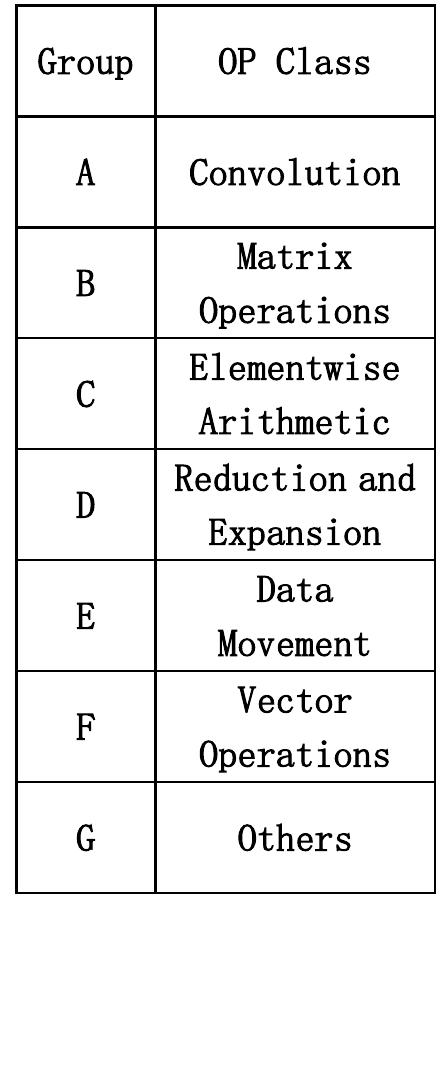}
\end{minipage}
\caption{Breakdown of execution time per operation type per deep learning model. Only operations that contribute to a least 1\% of the inference time are shown.}
 \label{fig:breakdown11}
\vspace{-5mm}
\end{figure*}

\section{Experimental Results}

\subsection{Roadmap}
%

In this section, we first quantify the computational characteristics of deep learning models. Next, we investigate how \quantization and
\pruning affect the model storage size and memory footprint. We then look at whether the reduction in the model size can translate into
faster inference time and lower power usage and energy consumption, as well as the implications of compression settings to the precision
metrics. Finally, we evaluate whether it is beneficial for combining both compress techniques.
\subsection{Model Computational Characteristics}
The first task of our experiments is to understand the computational characteristics for the deep learning models considered in this work.
Figure~\ref{fig:breakdown11} quantifies how different type of neural network operations contribute to the inference time across models. The
numbers are averaged across the test samples for each model -- 50K for image classifiers and 10K for \texttt{NMT}. To aid clarity, we group
the operations into seven classes, listed from A to G in the table on the right-hand side. Note that we only list operations that
contribute to at least 1\% of execution time.

 Each cell of the heatmap represents the
percentage that a specific type of operation contributes to the model inference time. As can be seen from the figure, a handful of
operations are collectively responsible for over 90\% of the model execution time. However, the types of operations that dominate the
inference time vary across networks. Unsurprisingly, \CNNs are indeed dominated by convolution, while fully-connected networks and \RNNs
depend heavily on matrix multiplications.

\begin{figure*}[!t]
\centering
\subfloat[][Model size]{\includegraphics[width=0.3\textwidth]{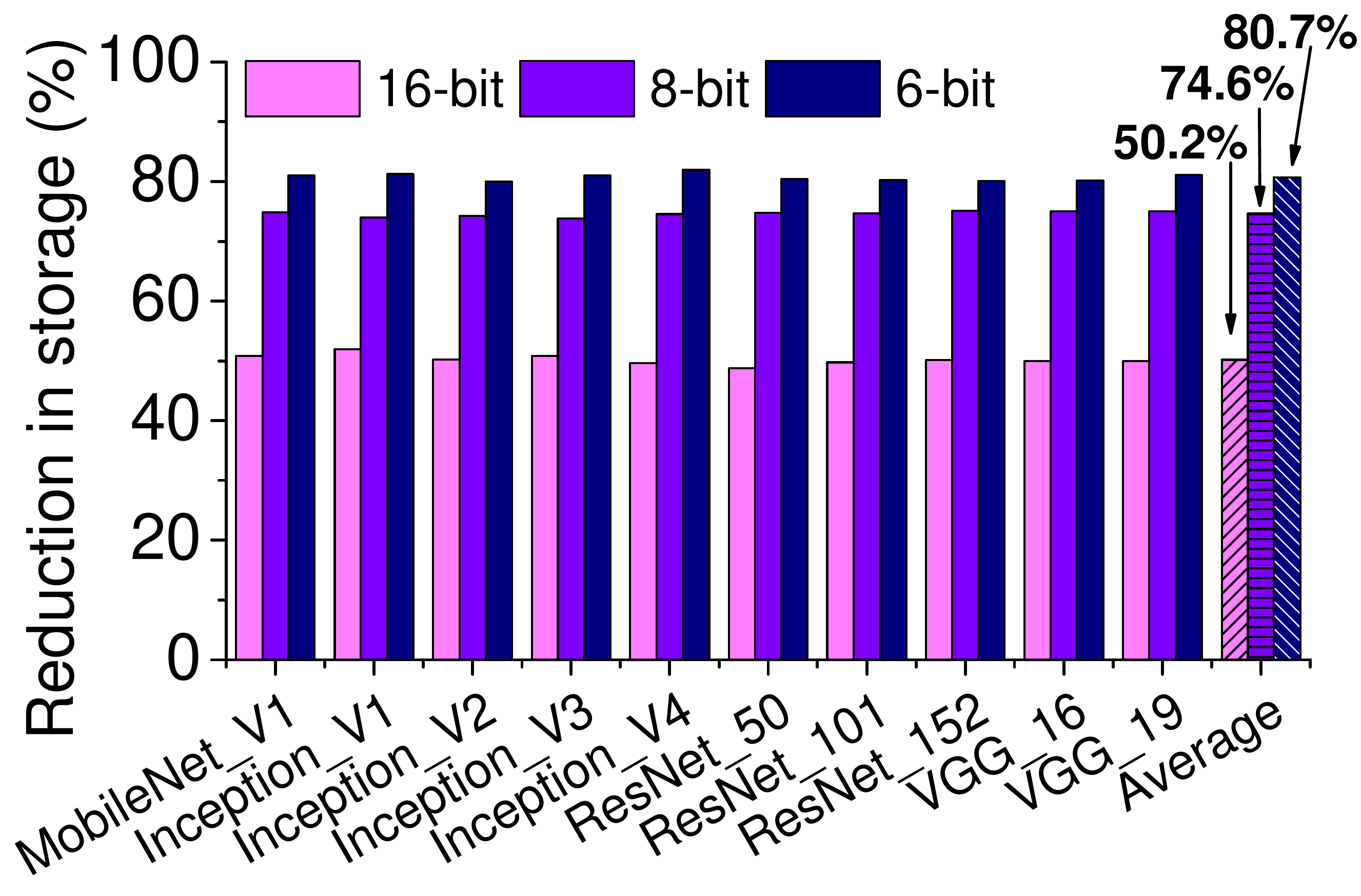}}
\hfill
\subfloat[][Inference time]{\includegraphics[width=0.31\textwidth]{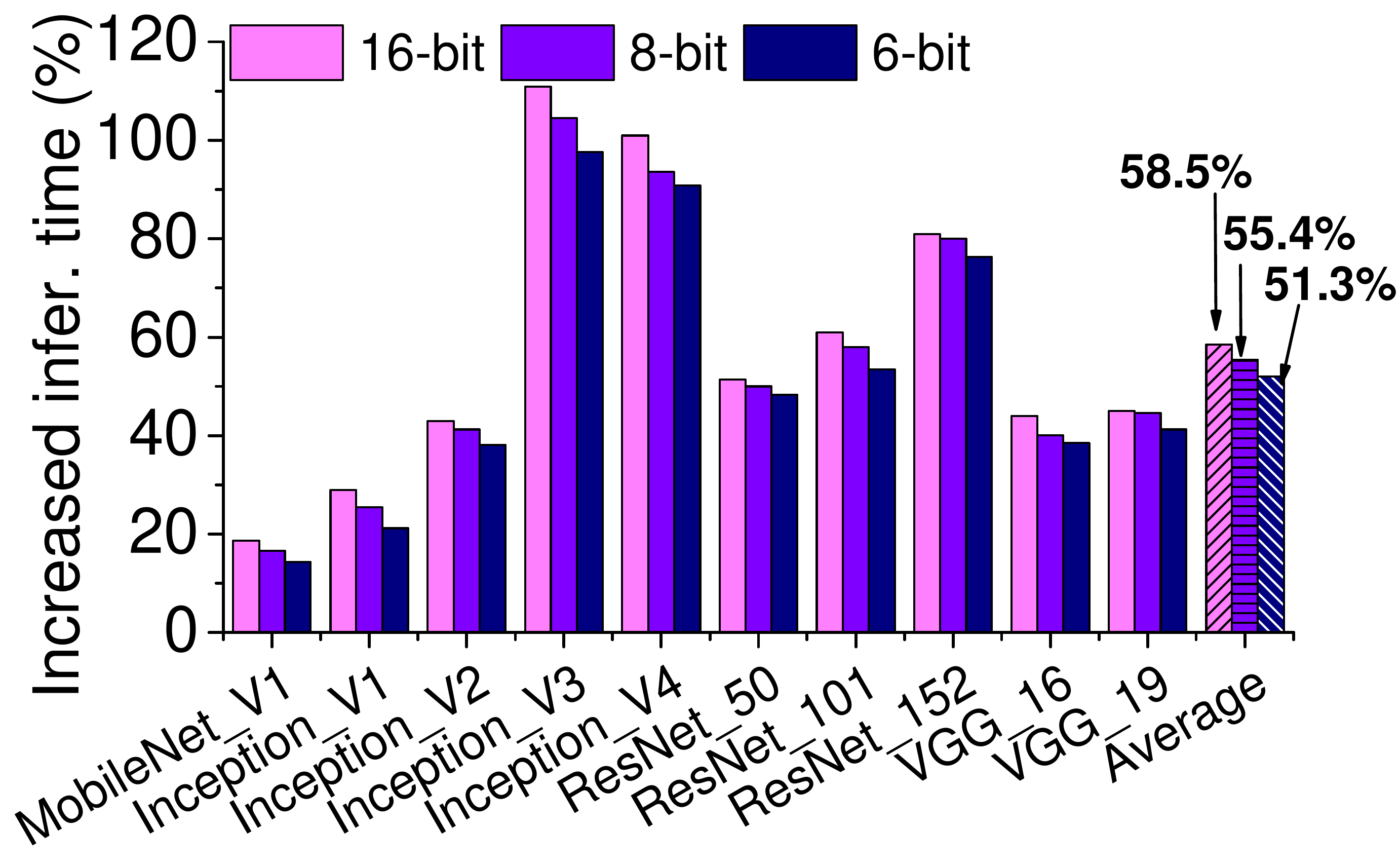}}
\hfill
\subfloat[][Accuracy]{\includegraphics[width=0.29\textwidth]{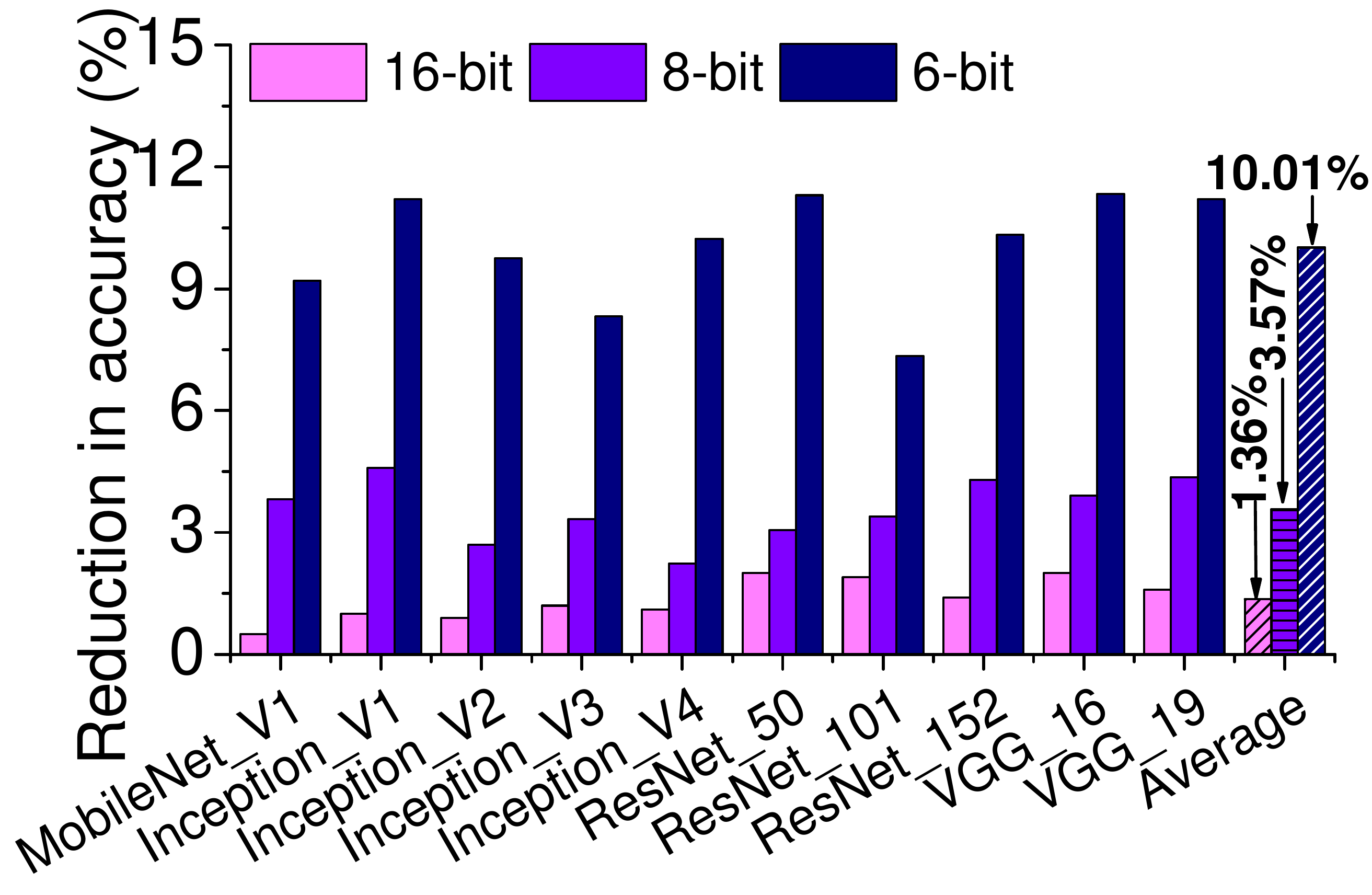}}
\hfill
\subfloat[][Power]{\includegraphics[width=0.3\textwidth]{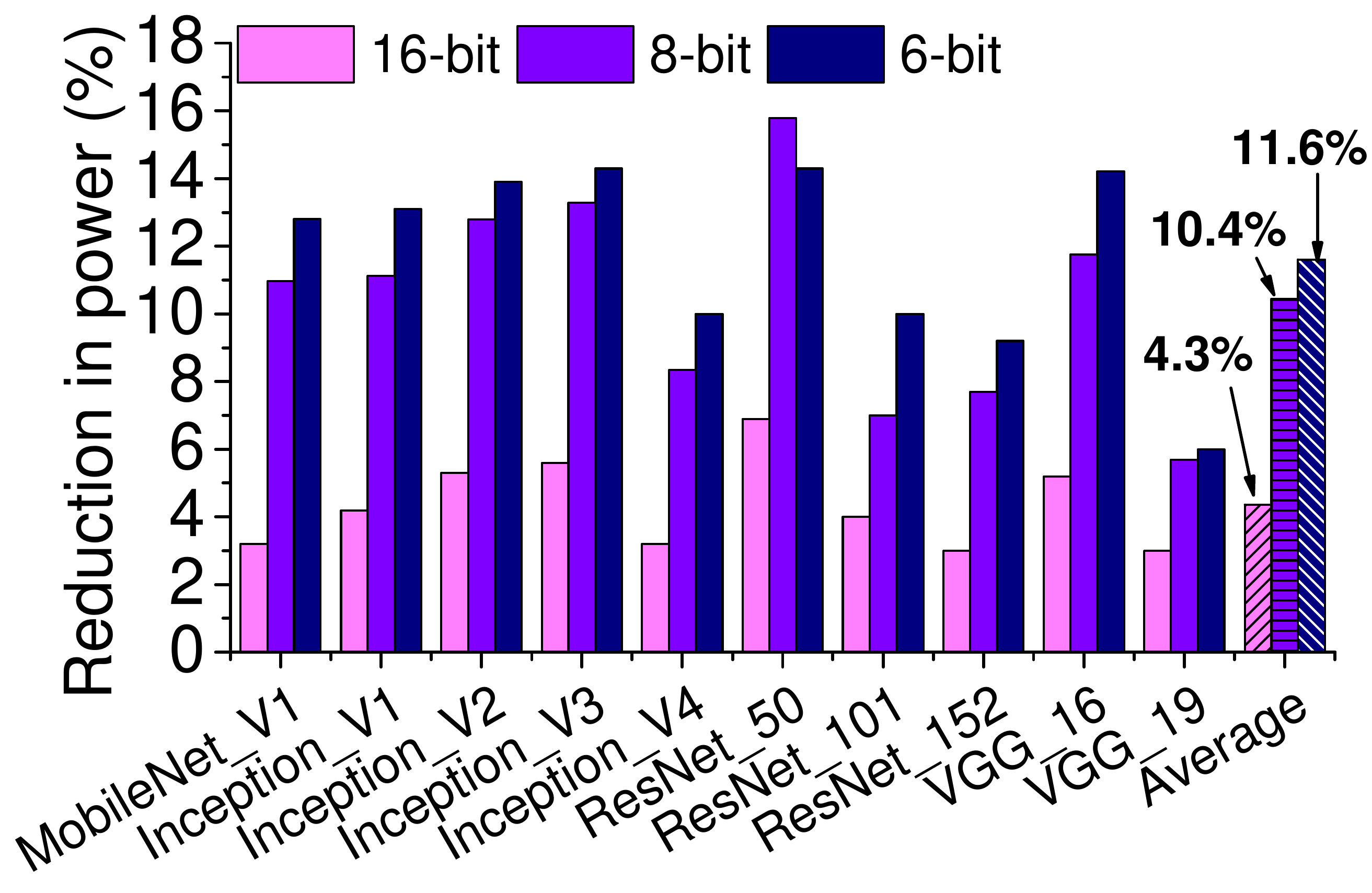}}
\hfill
\subfloat[][Energy consumption]{\includegraphics[width=0.31\textwidth]{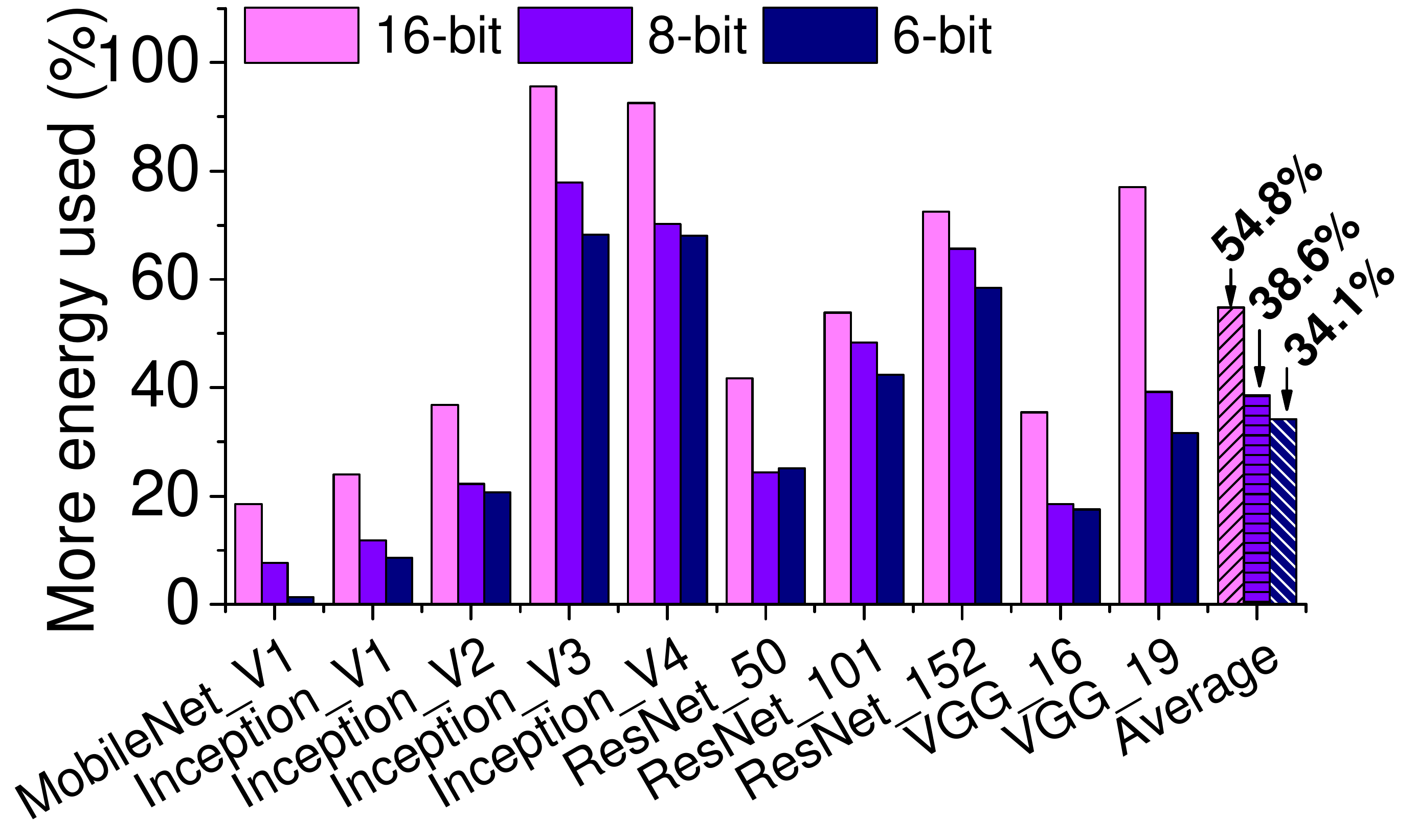}}
\hfill
\subfloat[][Precision, recall and F1-score]{\includegraphics[width=0.3\textwidth]{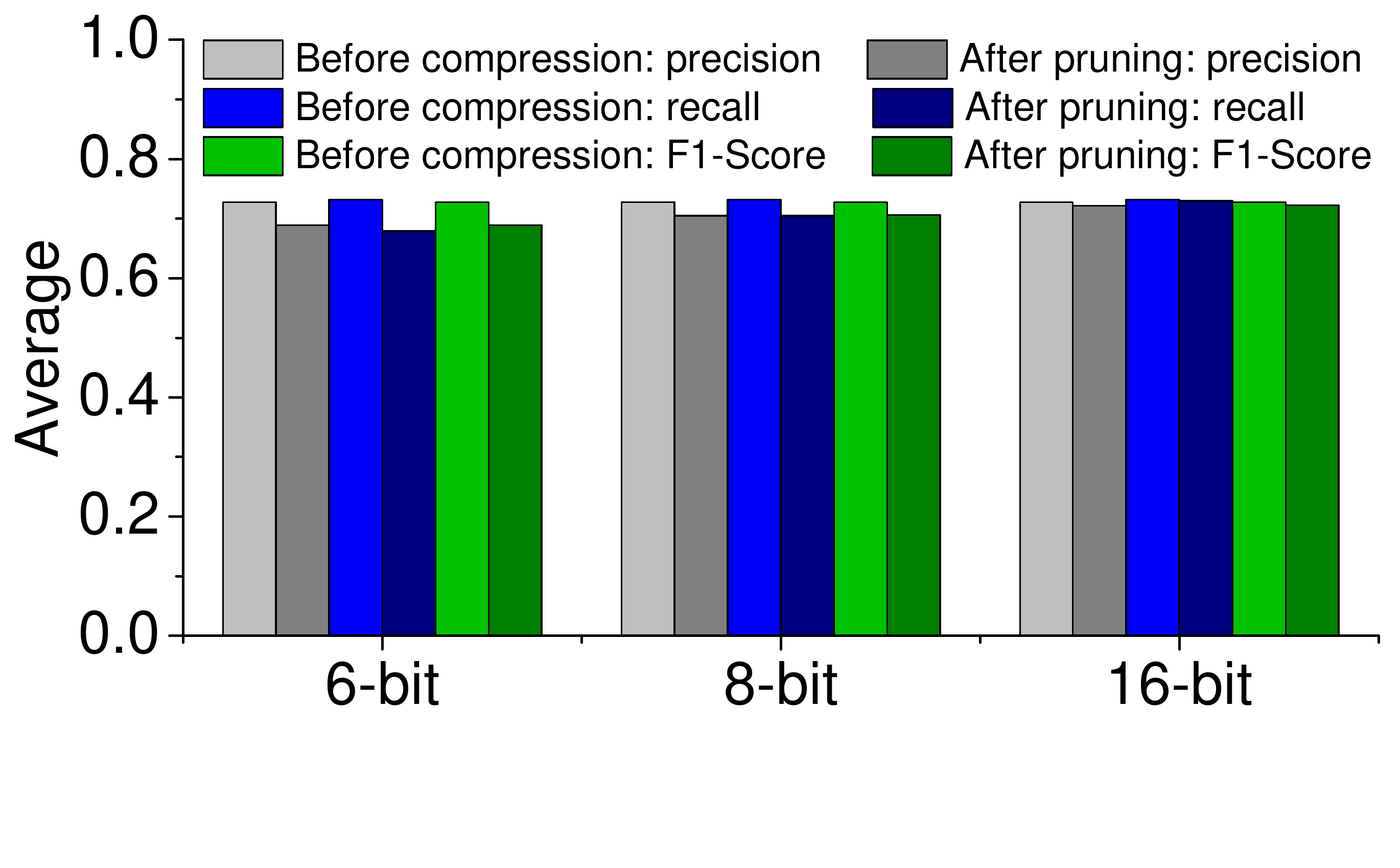}}
\hfill
\caption{The achieved model size (a) inference time (b) accuracy (c) power consumption (d)
energy consumption (e) and precision, recall and F1-score (e) before and after the compression by \quantization.
}
\label{fig:analy_quan}
\vspace{-5mm}
\end{figure*}

\begin{figure*}[!t]
\centering
\subfloat[][Model size]{\includegraphics[width=0.3\textwidth]{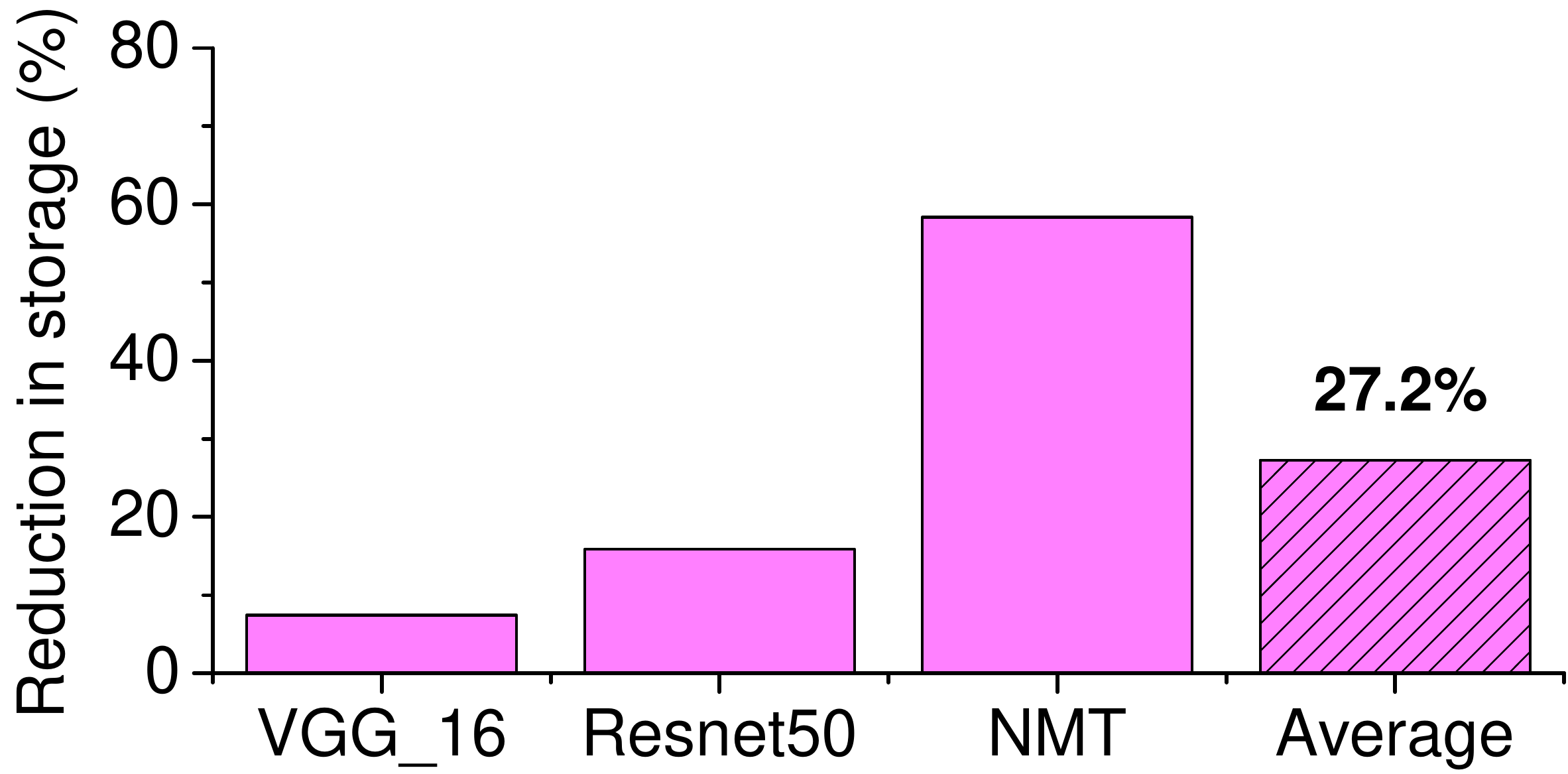}}
\hfill
\subfloat[][Inference time]{\includegraphics[width=0.27\textwidth]{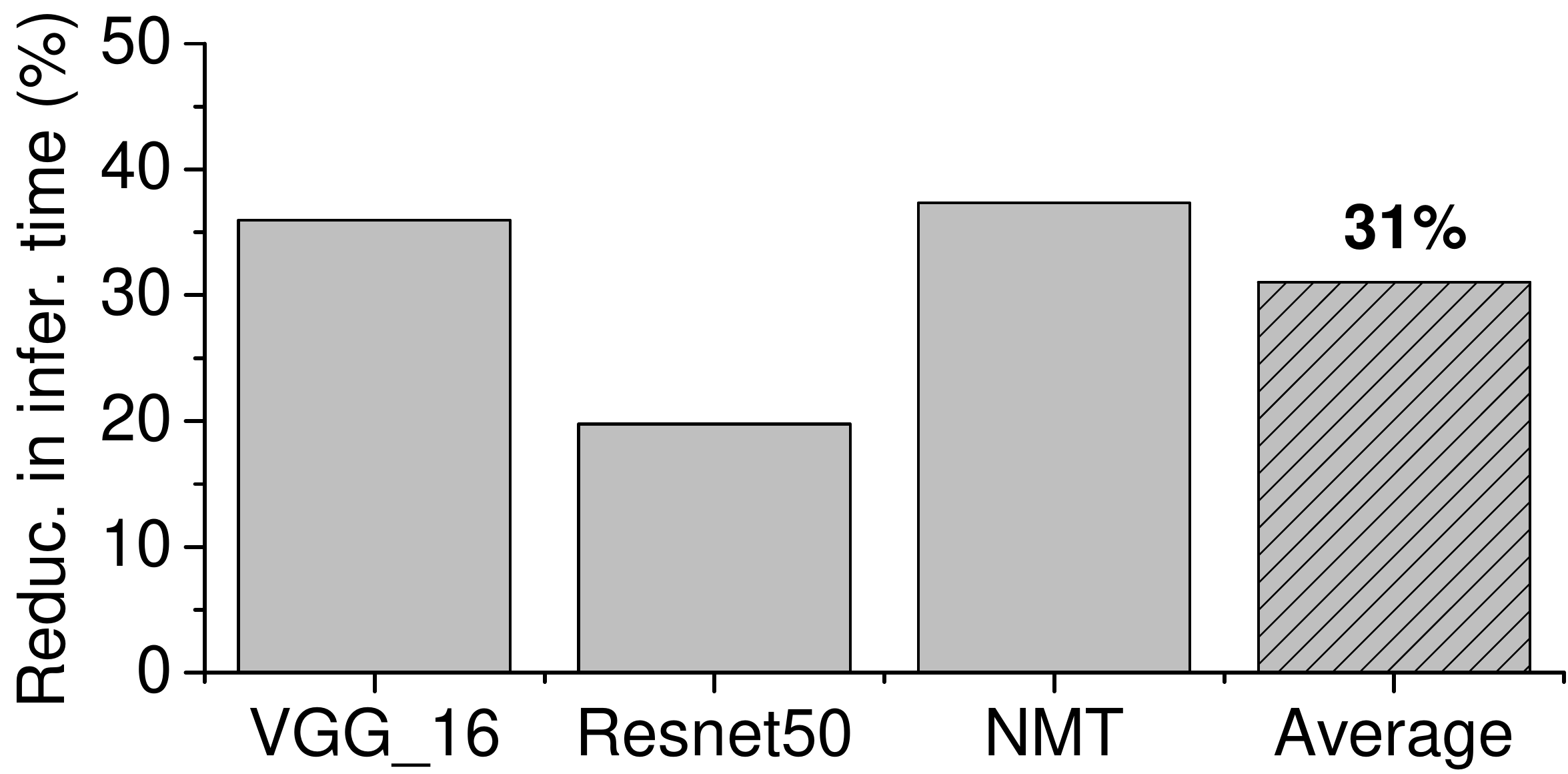}}
\hfill
\subfloat[][Accuracy]{\includegraphics[width=0.27\textwidth]{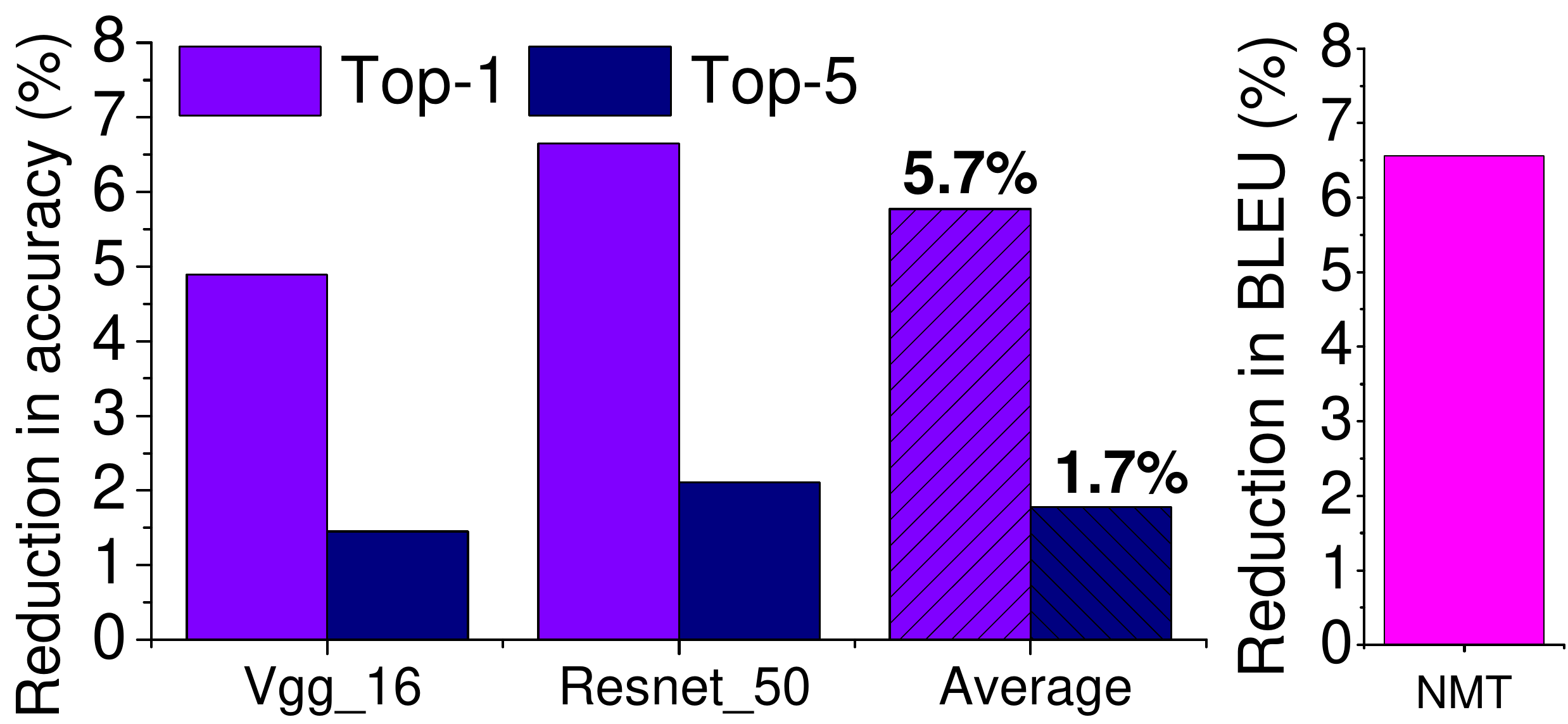}}
\hfill
\subfloat[][Power consumption]{\includegraphics[width=0.27\textwidth]{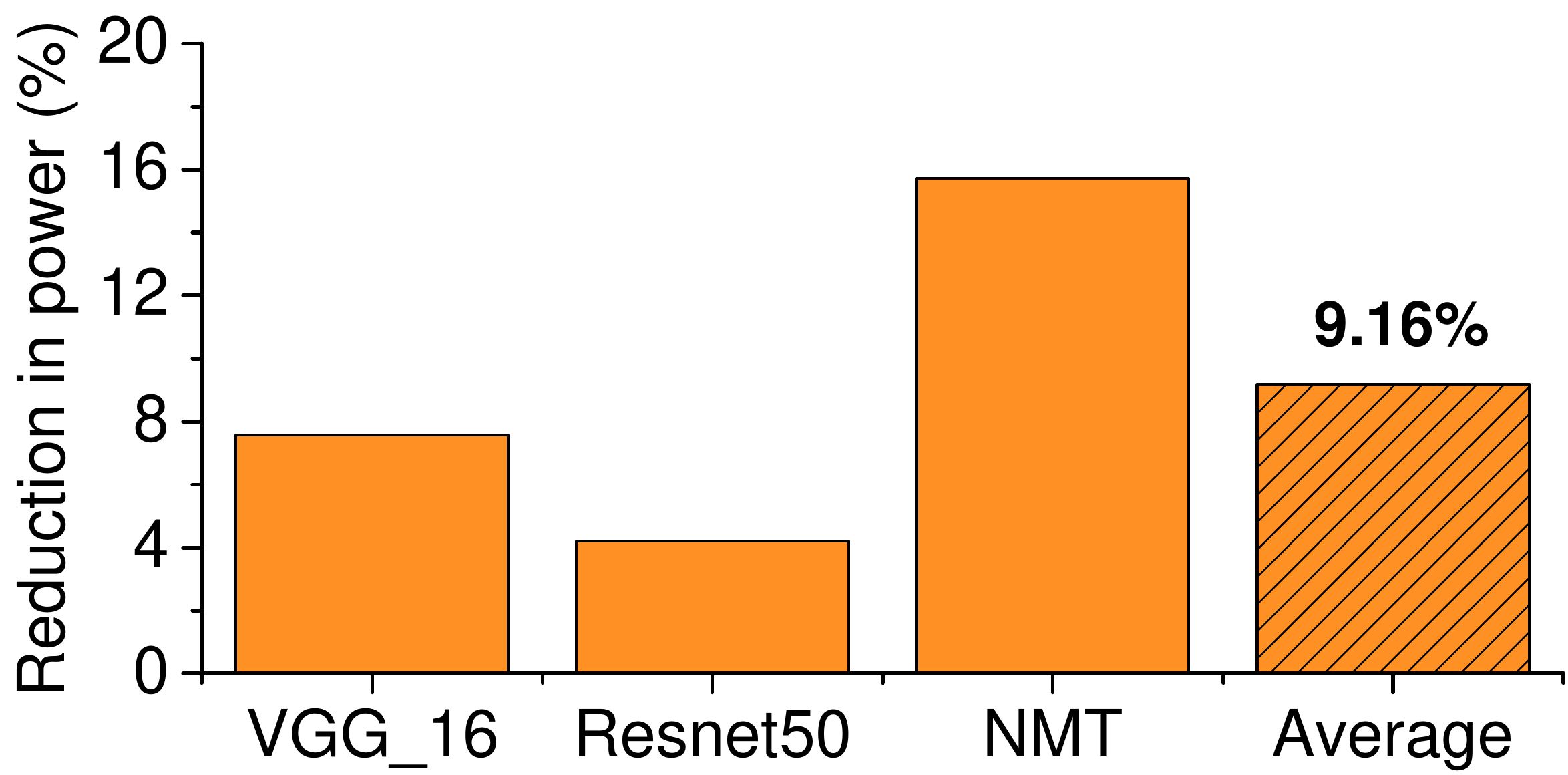}}
\hfill
\subfloat[][Energy consumption]{\includegraphics[width=0.27\textwidth]{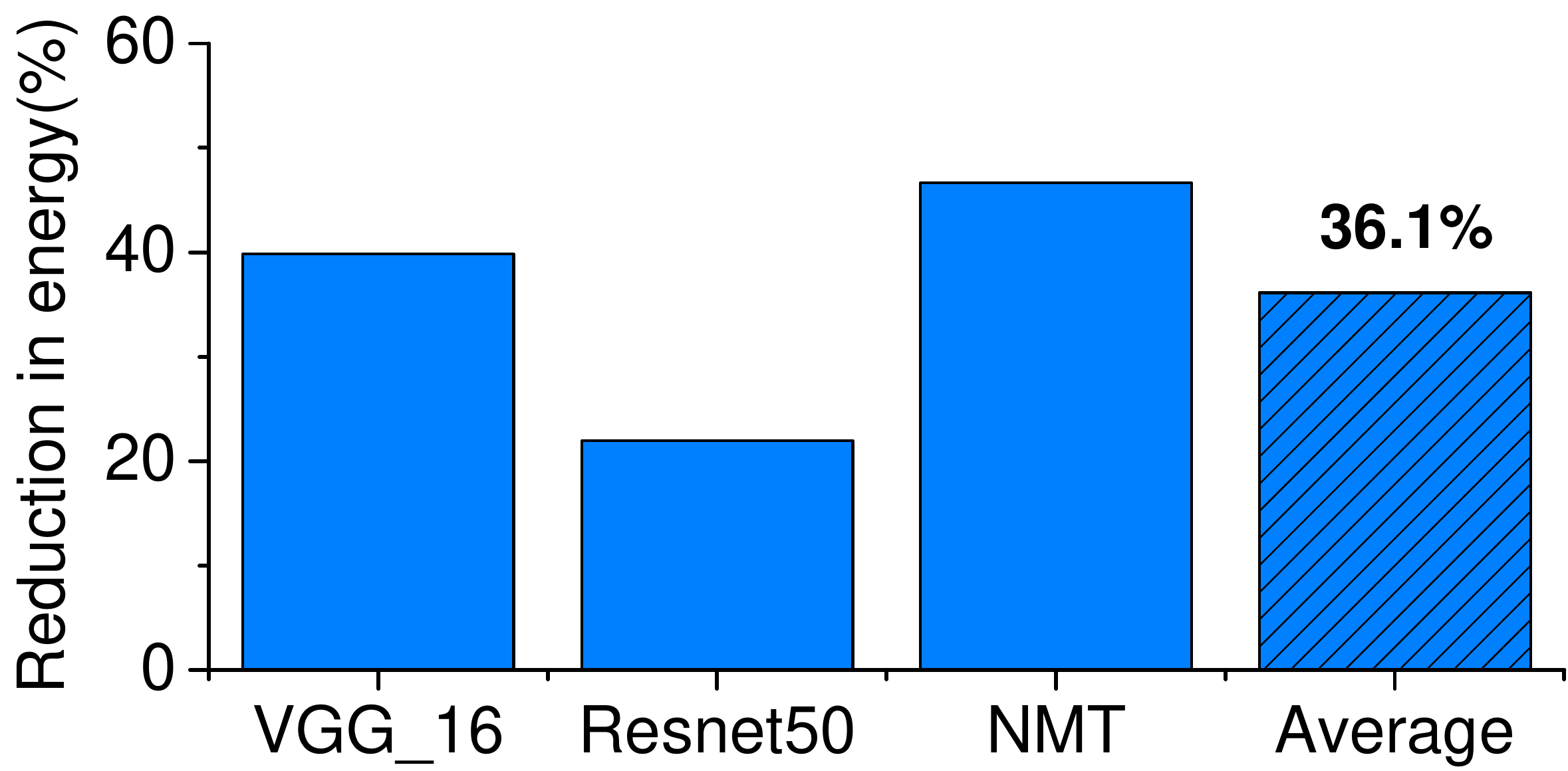}}
\hfill
\subfloat[][precision, recall and F1 score]{\includegraphics[width=0.27\textwidth]{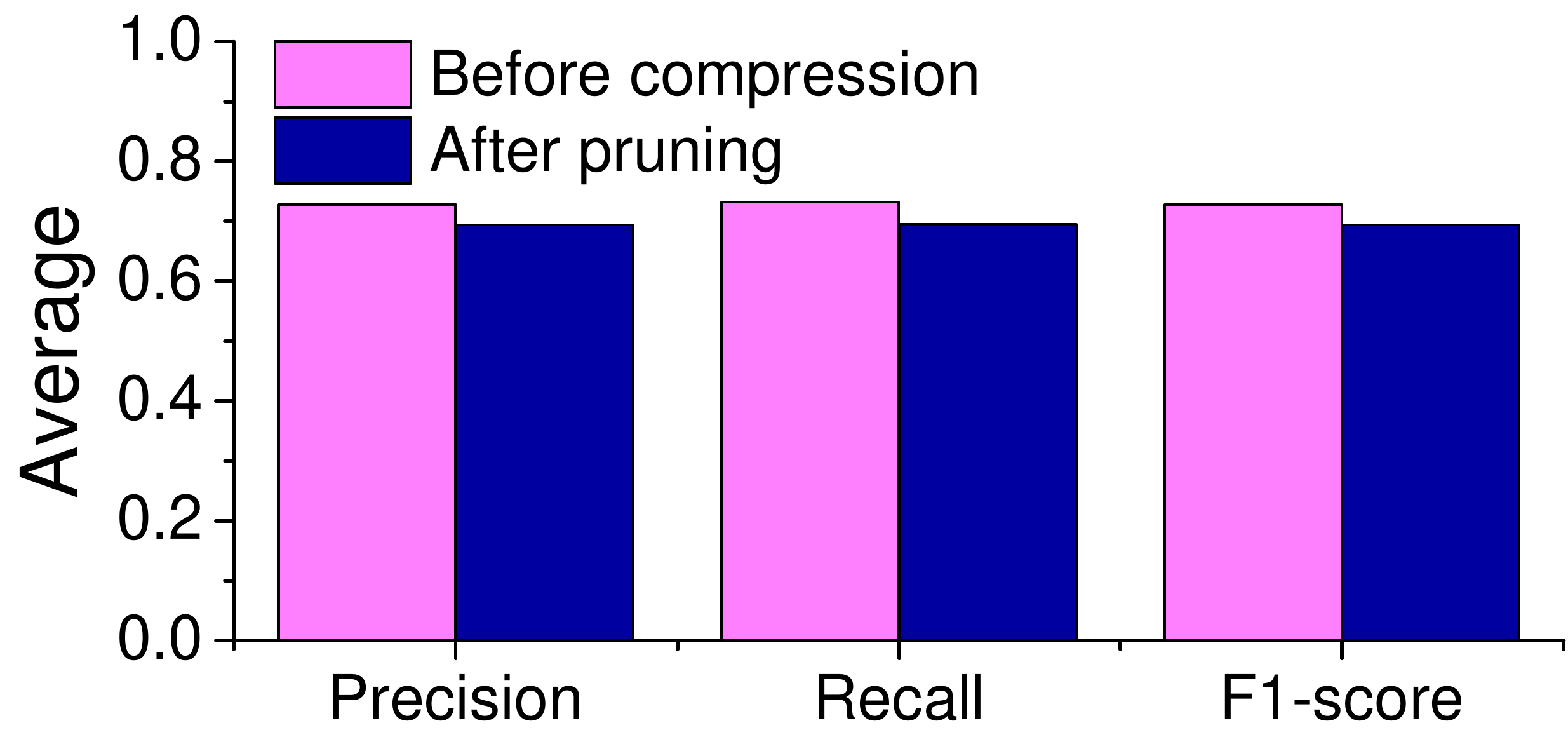}}
\hfill
\caption{The change of the model size (a), inference time (b), accuracy/BLEU (c), power (d), energy consumption (e), and accuracy (f) before and after applying \pruning.} \label{fig:analy_prun}
\vspace{-5mm}
\end{figure*}

\begin{figure*}[!t]
\centering
\subfloat[][VGG\_16]{\includegraphics[width=0.28\textwidth]{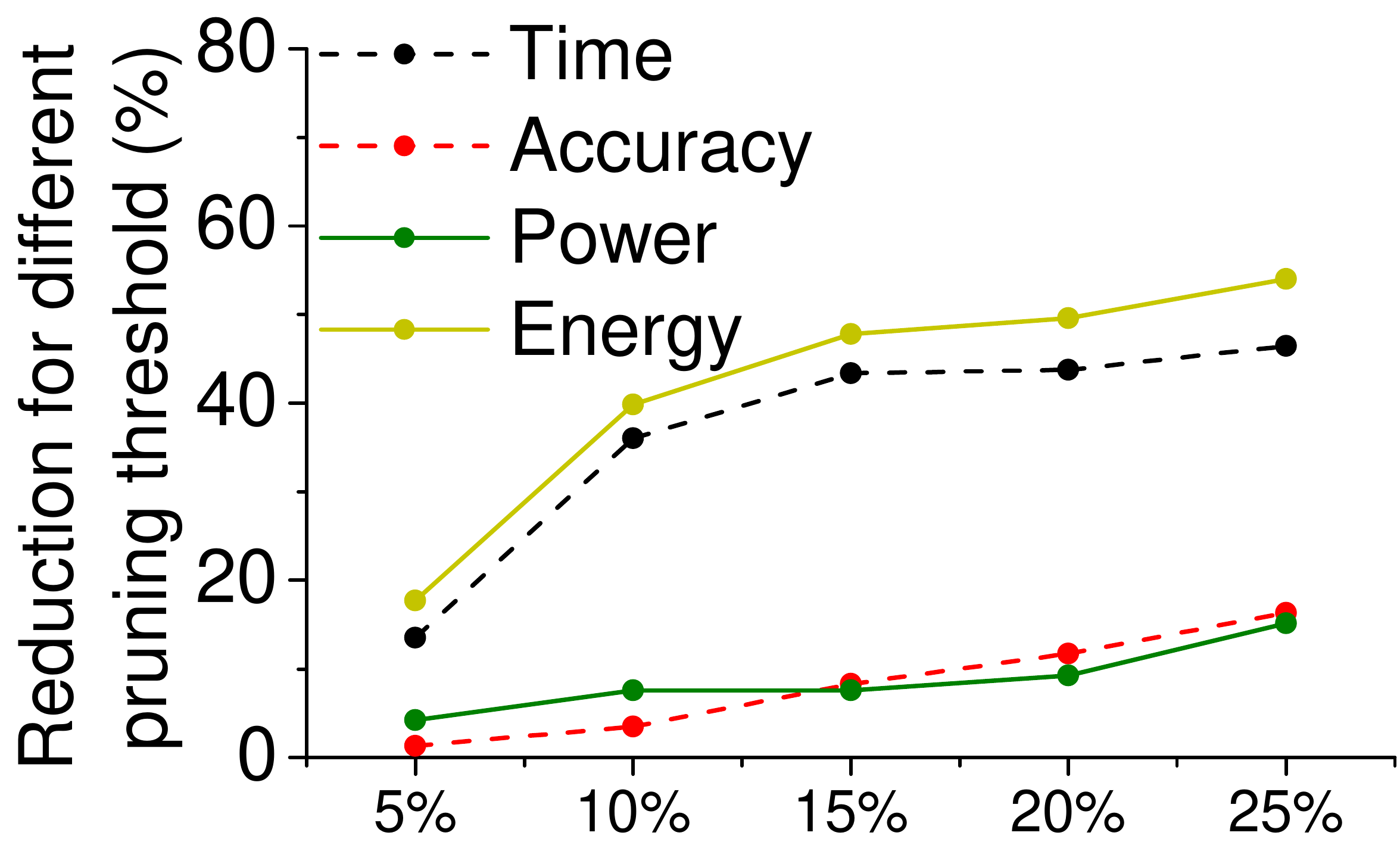}}
\hfill
\subfloat[][ResNet\_50]{\includegraphics[width=0.28\textwidth]{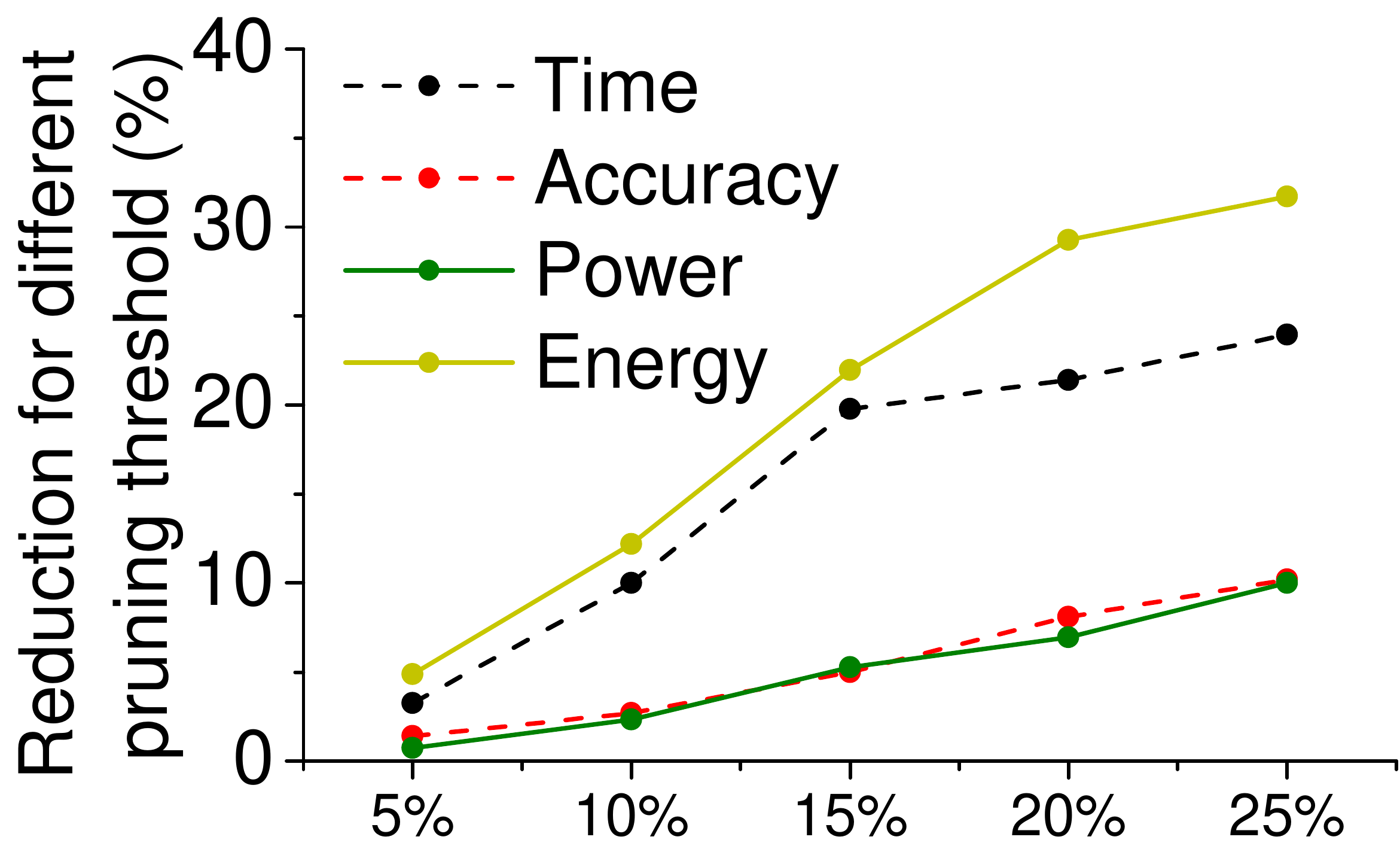}}
\hfill
\subfloat[][NMT]{\includegraphics[width=0.28\textwidth]{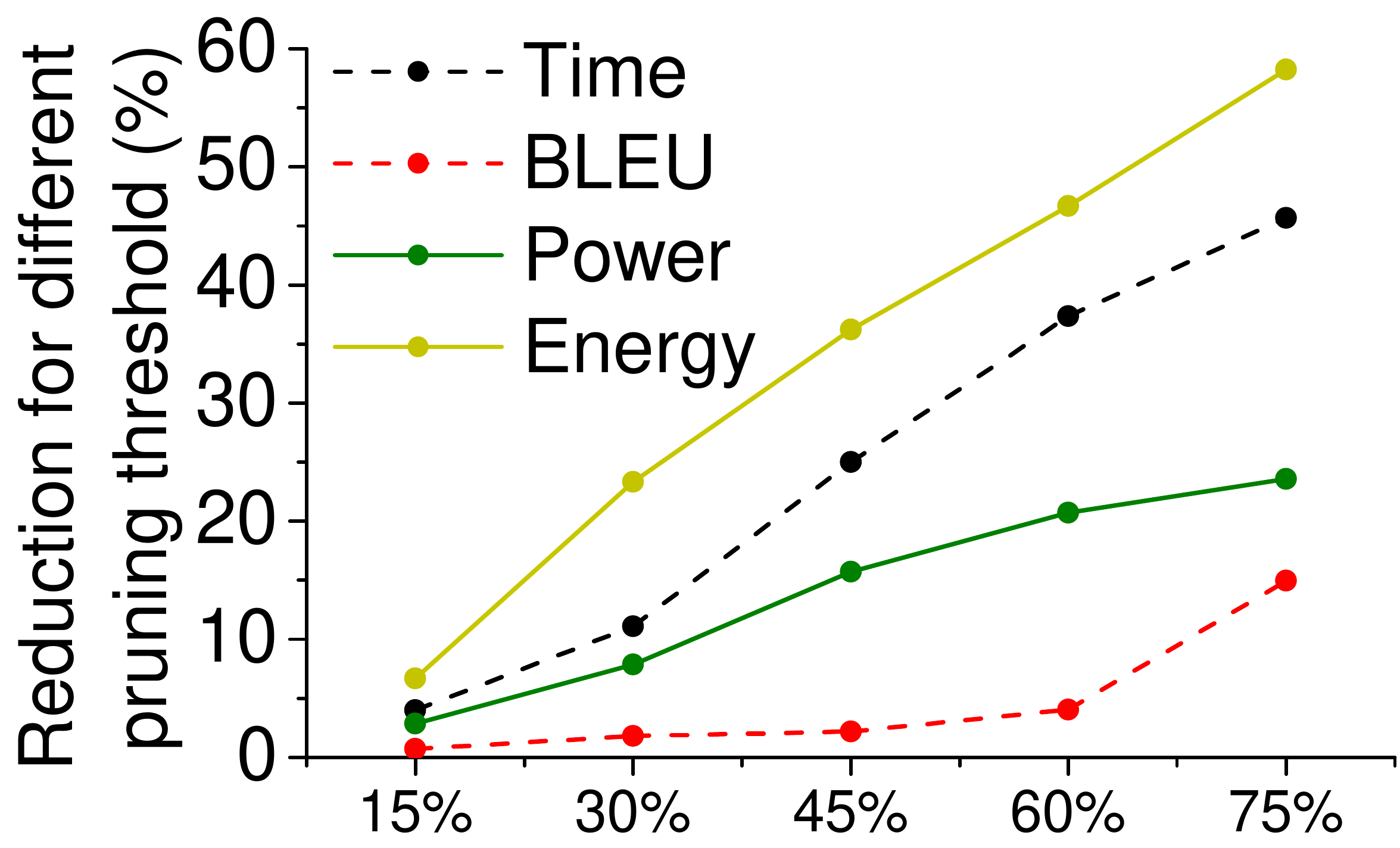}}
\hfill
\caption{The resulting inference time, accuracy, and power and energy consumption
for VGG\_16 (a), ResNet\_50 (b) and \texttt{NMT} (c) when using different pruning thresholds.
The x-axis shows the percentage of pruning in model size. }
\label{fig:threshold}
\vspace{-7mm}
\end{figure*}

\subsection{Impact on the Model Storage Size\label{sec:ms}}
Reducing the model storage size is crucial for storage-constrained devices. A smaller model size also translates to smaller runtime memory
footprint of less RAM space usage. Figures~\ref{fig:analy_quan}, \ref{fig:analy_prun} and \ref{fig:threshold} illustrate how the different
compression techniques and parameters affect the resulting model size.

As can be seen from Figure~\ref{fig:analy_quan}a, data quantization can significantly reduce the model storage size, leading to an average
reduction of 50.2\% when using a 16-bit representation and up to 80.7\% when using a 6-bit representation. The reduction in the storage
size is consistent across neural networks as the size of a network is dominated by its weights.

From Figure~\ref{fig:analy_prun}a, we see that by removing some of the neurons of a network, \pruning can also reduce the model size,
although the gain is smaller than \quantization if we want to keep the accuracy degradation within 5\%. On average, \pruning reduces the
model size by 27.2\% (49.26 MB). An interesting observation is that, \pruning is particularly effective for obtaining a compact model for
\texttt{NMT}, an \RNN, with a reduction of 60\% on the model storage size. This is because there are many repetitive neurons (or cells) in
an \RNN due to the natural of the network architecture. As we will discuss later, \pruning only causes a minor degradation in the
prediction accuracy for \texttt{NMT}. This suggests that \pruning can be an effective compression technique for \RNNs.

In Figure~\ref{fig:threshold}, we compare the resulting performance after using different pruning thresholds from 5\% to 75\% on two \CNN
and a \RNN models. Increasing the pruning percentage of model size provides  more opportunities for \pruning to remove more neurons to
improve the inference time and other metrics. The improvement reaches a plateau at 15\% of reduction for \CNNs, but for \texttt{NMT}, a
\RNN, the reduction of inference time increases as we remove more neurons. This diagram reinforces our findings that \pruning is more
effective on \RNNs than \CNNs. For the rest discussions of this paper, we use a 5\% \pruning threshold.


\subsection{Memory Footprint}

\begin{figure}[!t]
\centering
\subfloat[][\quantization]{\includegraphics[width=0.25\textwidth]{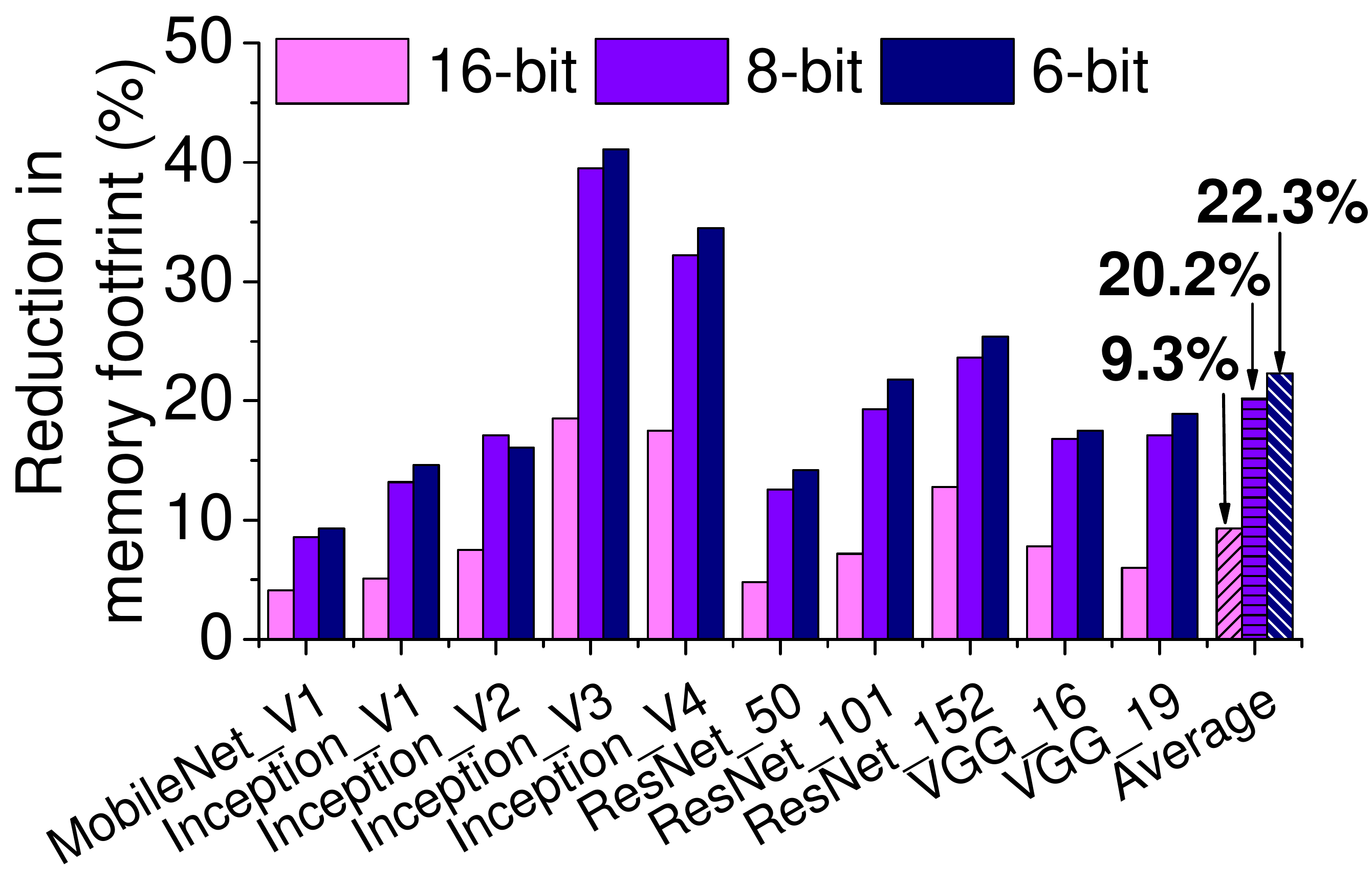}}
\hfill
\hspace{-2mm}
\subfloat[][\pruning]{\includegraphics[width=0.23\textwidth]{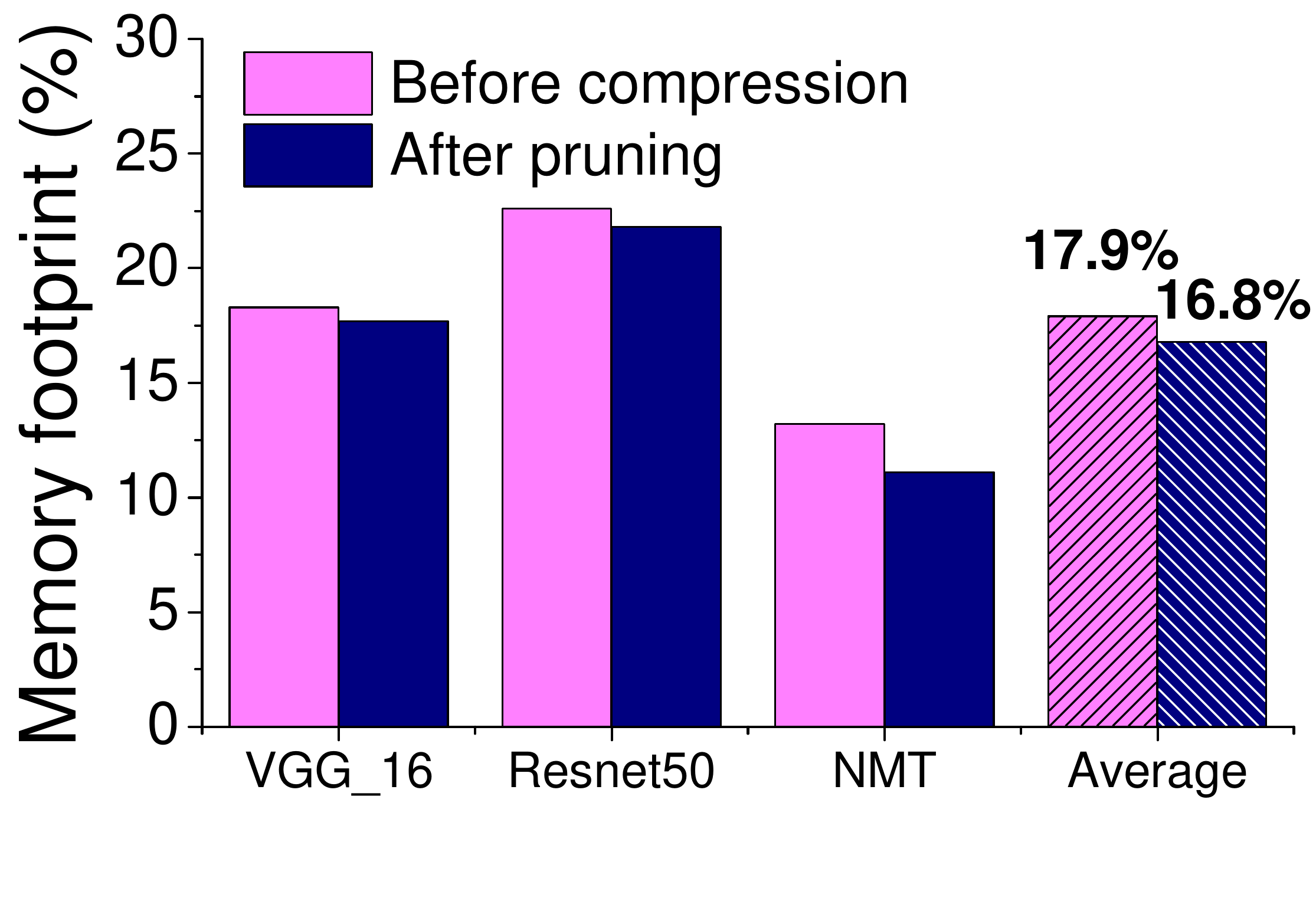}}
\hfill

\caption{Memory footprint before and after applying \quantization(a) and \pruning (b).} \label{fig:footprint}
\vspace{-6mm}
\end{figure}

Figure~\ref{fig:footprint} compares the runtime memory footprint consumed by a compressed model. Quantization reduces the model memory
footprint by 17.2\% on average. For example, an 8-bit representation gives a memory footprint saving from 20.02\% to 15.97\% across
networks, with an averaged reduction of 20.2\% (up to 40\%). In general, the smaller the model is, the less memory footprint it will
consume. As an example, a 6-bit representation uses 2.6\% and 13.6\% less memory compared to an 8-bit and a 16-bit counterparts,
respectively.

Figure~\ref{fig:footprint} suggests that  \pruning offers little help in reducing the model memory footprint. On average, it gives a 6.1\%
reduction of runtime memory footprint. This is because that the network weights still domain the memory resource consumption, and \pruning
is less effective compared to \dquantization for reducing the overhead of network weights.

\subsection{Impact on Inference Time\label{sec:time}}
\begin{figure}
\begin{center}
\includegraphics[width=0.35\textwidth]{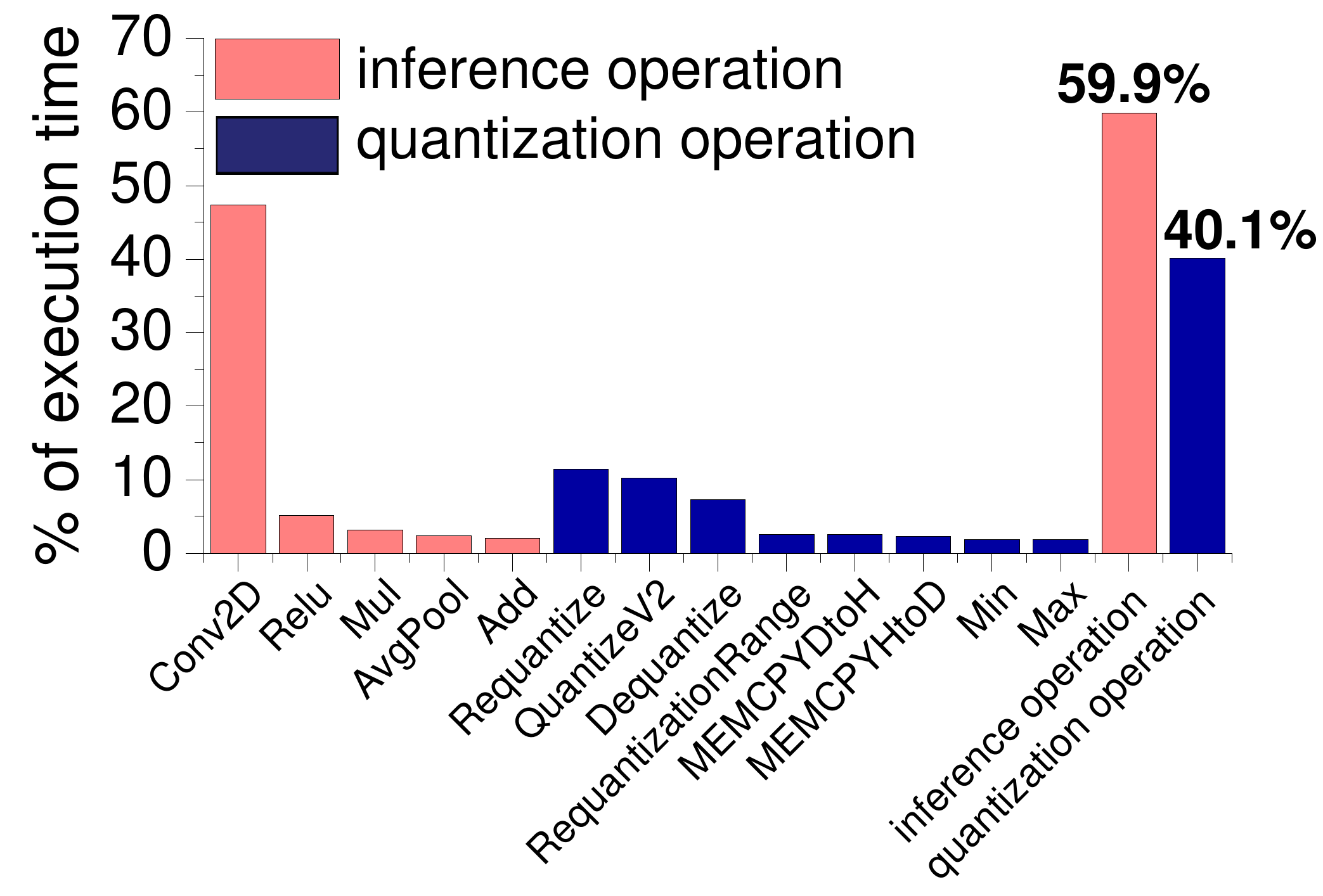}
\end{center}
\vspace{-2mm}
\caption{Breakdown of the averaged execution time per operation type, averaging across the quantized models.}
\vspace{-6mm}
\label{fig:breakdown}
\end{figure}

 Figure~\ref{fig:analy_quan}b compares the inference time when using different bit
widths to represent a 32-bit floating number for neural network weights. Intuitively, a smaller model should run faster. However, data
quantization does not shorten the inference time but prolongs it. The reasons are described as follows. Data quantization can speedup the
computation (i.e., matrix multiplications) performed on some of the input data by avoiding expensive floating point arithmetics and
enabling SIMD vectorization by using a compact data representation. However, we found that the overhead of the de-quantization process
during inference can outweigh its benefit. Besides the general inference operation, a data quantization and de-quantization function has to
be added into the compressed model. Inference performed on a quantized model accounts for 59.9\% of its running time. The de-quantization
functions converts input values back to a 32-bit representation on some of the layers (primarily the output layer) in order to recover the
loss in precision. As can be seen from Figure~\ref{fig:breakdown}, this process could be expensive, contributing to 30\% to 50\% of the
inference time.

Using fewer bits for representation can reduce the overhead of de-quantization. For example, using a 6-bit representation is 1.05x and
1.03x faster than using a 16-bit and a 8-bit representations, respectively. However, as we will demonstrate later when discussing
Figure~\ref{fig:analy_quan}c, using fewer bits has the drawback of causing larger degradation in the prediction accuracy. Hence, one must
carefully find a balance between the storage size, inference time, and prediction accuracy when applying data quantification.

We also find that the percentage of increased inference time depends on the neural network structure. Applying data quantization to
\texttt{Inception}, the most complex network in our \CNN tested set, will double the inference time. By contrast, data quantization only
leads to a 20\% increase in inference time for \texttt{Mobilenet}, a compact model. This observation suggests that data quantization may be
beneficial for simple neural networks on resource-constrained devices.

In contrast to \quantization, Figure~\ref{fig:analy_prun}b shows that \pruning leads to faster inference time across evaluated networks,
because there is no extra overhead added to a pruned network. We see that the inference time of \texttt{VGG\_16} and \texttt{NMT} can
benefit from this technique, with an reduction of 38\%. Overall, the average inference time is reduced by 31\%. This suggests that while
\pruning is less effective in reducing the model size (see Section~\ref{sec:ms}), it is useful in achieving a faster inference time.

\subsection{Impact on the Power and Energy Consumption}
Power and energy consumption are two limiting factors on battery-powered devices. As we can see from Figure~\ref{fig:analy_quan}d,
\quantization decreases the peak power usage for inferencing with an average value of 4.3\%, 10.4\% and 11.6\%, respectively when using a
6-bit, an 8-bit and a 16-bit representations. While \quantization reduces the peak power the device draws, the increased inference time
leads to more energy consumption (which is a product of inference time $\times$ instantaneous power) by at least 34.1\% and up to 54.8\%
(see Figure~\ref{fig:analy_quan}e). This means that although \quantization allows one to reduces supplied power and voltage, it can lead to
a short battery life.

Figure~\ref{fig:analy_prun}e quantifies the reduced energy consumption by applying \pruning. Note that it saves over 40\%, 15\% and 50\%
energy consumption for \texttt{VGG\_16}, \texttt{ResNet\_50} and \texttt{NMT} respectively. Despite that achieving only minor decreases in
the peak power usage compared to \quantization (9.16\% vs 34.1\% to 51.1\%), the faster inference time of \pruning allows it to reduce the
energy consumption. The results indicate \pruning is useful for reducing the overall energy consumption of the system, but \quantization
can be employed to support a low power system.

\subsection{Impact on The Prediction Accuracy}
Accuracy is obviously important for any predictive model because a small and faster model is not useful if it gives wrong predictions all
the time.

Results in Figure~\ref{fig:analy_quan}c compare how the prediction accuracy is affected by model compression. We see that the sweat spot of
\quantization depends on the neural network structure. An 16-bit representation keeps the most information of the original model and thus
leads to little reduction in the prediction accuracy, on average  1.36\%.  Using an 8-bit representation would lead on average 3.57\%
decrease in the accuracy, while using a 6-bit representation will lead to a significantly larger reduction of 10\% in  accuracy. We also
observe that some networks are more robust to \quantization. For example, while a 6-bit representation leads to less than 10\% decrease in
accuracy for \texttt{ResNet\_101}, it cause a 12\% drop in accuracy for \texttt{ResNet\_50}. This is because a more complex network (i.e.,
\texttt{ResNet\_101} in this case) is more resilient to the weight errors compared to a network (i.e., \texttt{ResNet\_50} in this case)
with a smaller number of layers and neurons. Our findings suggest the need for having an adaptive scheme to choose the optimal
\dquantization parameter for given constraints.

For \pruning, Figure~\ref{fig:analy_prun}c compares the reduction in the top-1 and the top-5 scores for \texttt{VGG\_16} and
\texttt{ResNet\_50}. We also show the BLEU value for \texttt{NMT}. Overall, \pruning reduces the accuracy of the two \CNN models with by
5.7\% and 1.7\% respectively for the top-1 and the top-5 scores. It has little negative impact on \texttt{NMT} where we only observe an
averaged loss of 2.1\% for BLEU. When taking into consideration that \pruning can significantly reduce the model size for \texttt{NMT}
(Section~\ref{sec:ms}), our results suggest that \pruning is particularly effective for \RNNs.

\subsection{Precision, Recall and F1-Score}

Figures~\ref{fig:analy_quan}f and \ref{fig:analy_prun}f show other three performance metrics of a \emph{classification} model after
applying \quantization and \pruning. We can see that, the decrease in performance after compression is  less than 3\% for precision, recall
and the F1-score. For \quantization, the 16-bit representation outperforms the other two bits width representations. Specifically,  a
16-bit representation gives the highest overall precision, which in turns leads to the best F1-score. High precision can reduce false
positive, which is important for certain domains like video surveillance because it can reduce the human involvement for inspecting false
positive predictions.

\begin{figure*}[!t]
\centering
\subfloat[][Inference time vs accuracy]{\includegraphics[width=0.3\textwidth]{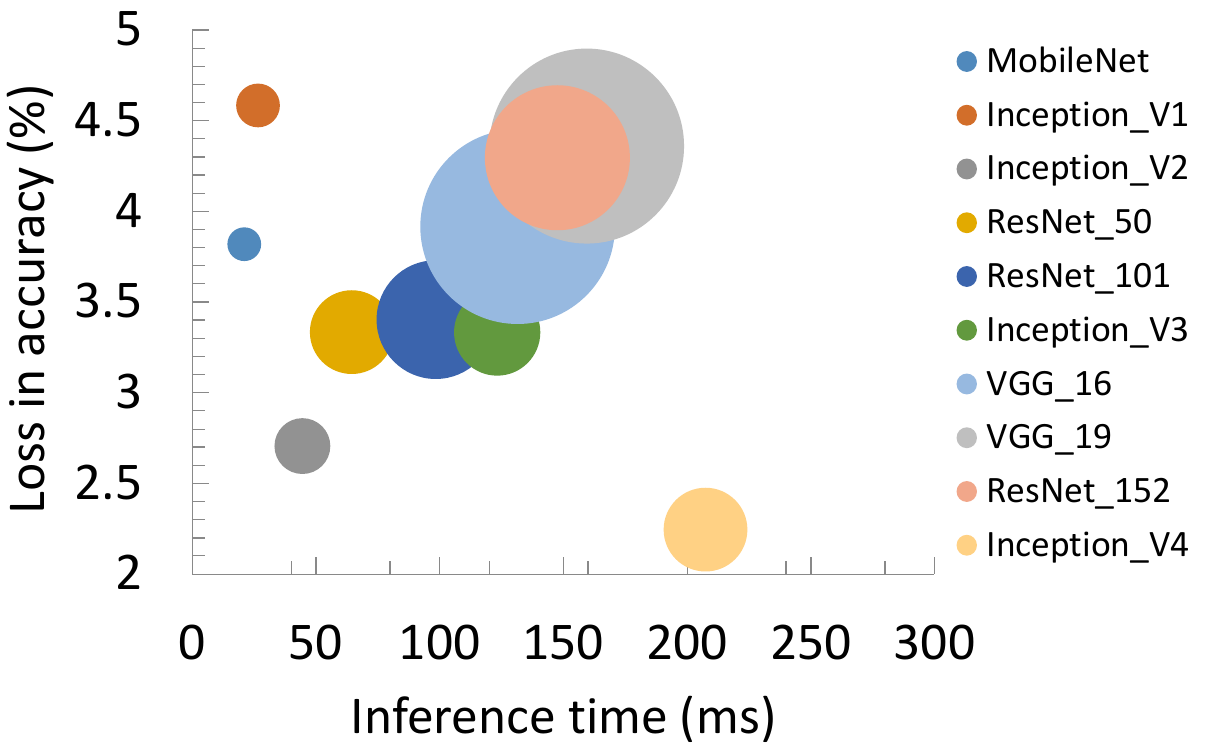}}
\hfill
\subfloat[][Energy vs accuracy]{\includegraphics[width=0.3\textwidth]{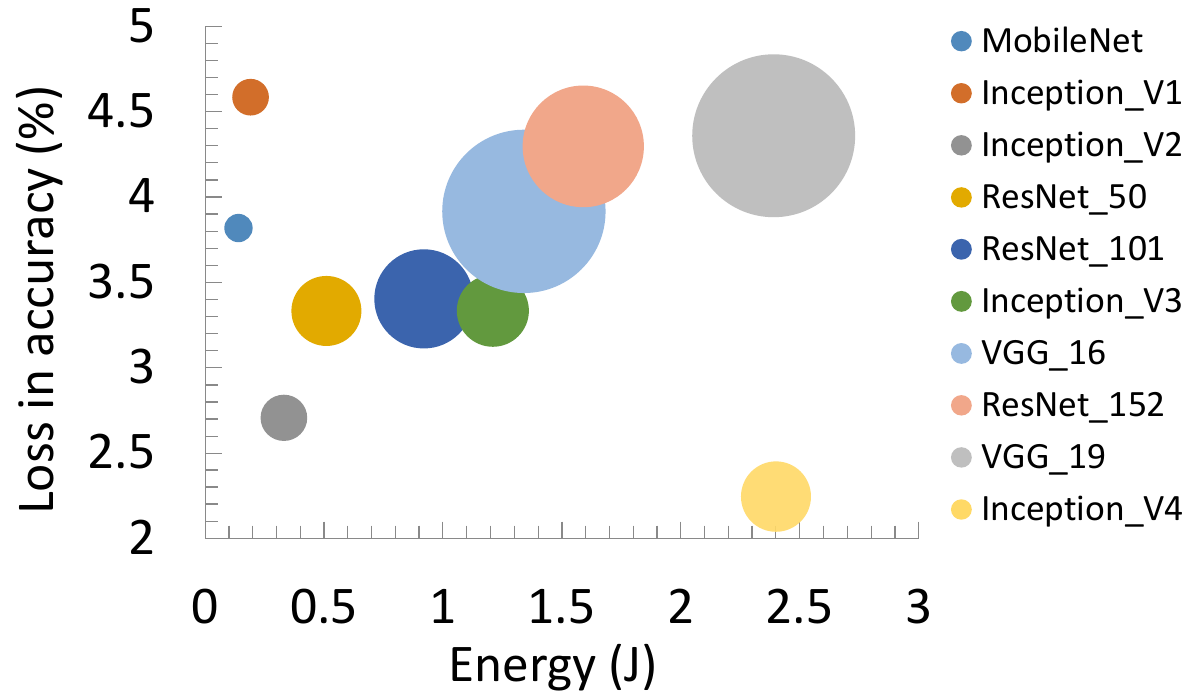}}
\hfill
\subfloat[][Compressed model size vs accuracy]{\includegraphics[width=0.3\textwidth]{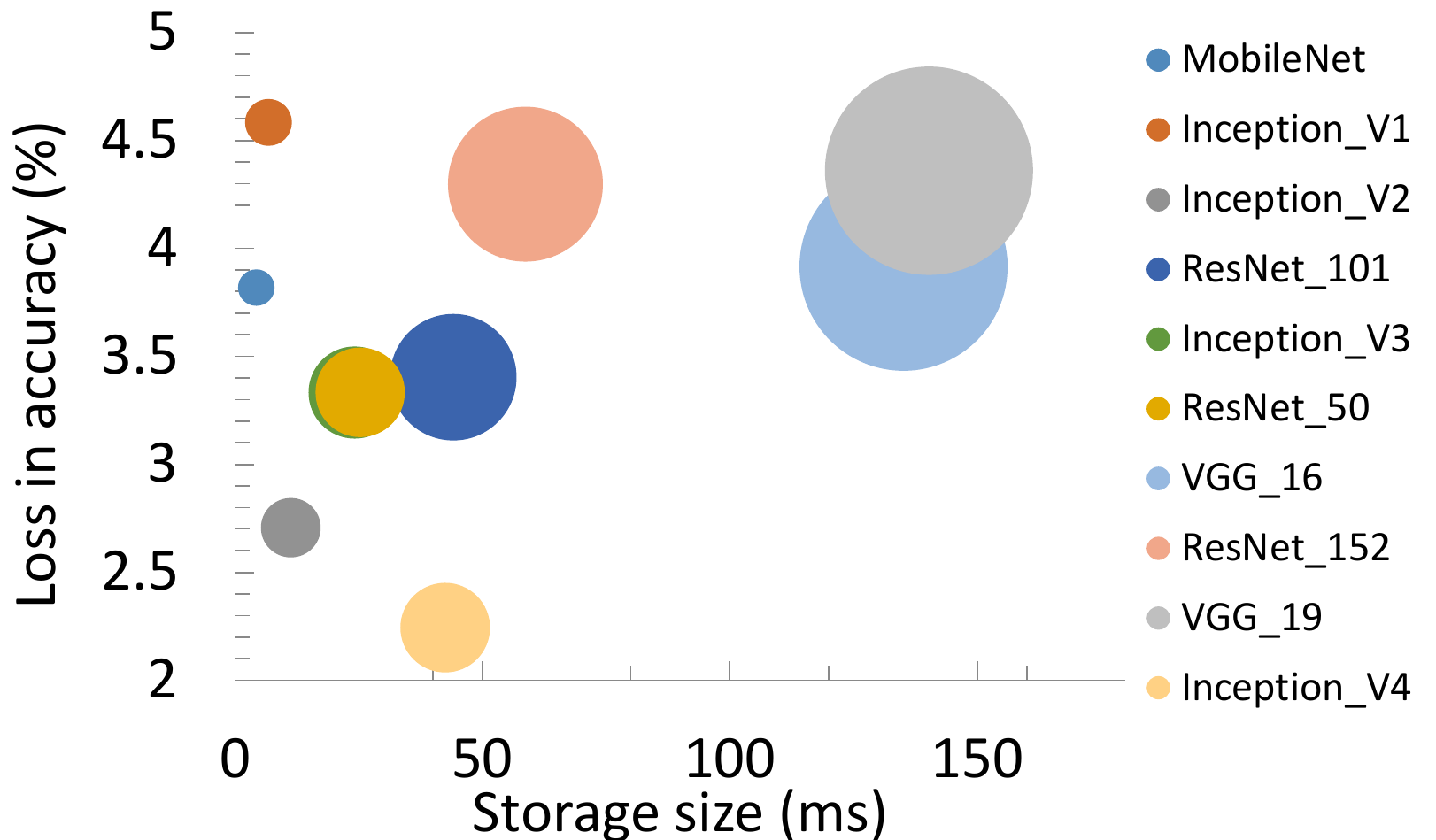}}
\caption{The \quantization effects on different sized models. The larger a bubble is, the more parameters the corresponding model has (see
Table~\ref{tab:workload}).}
\label{fig:bubblequantization}
\vspace{-7mm}
\end{figure*}

\begin{figure*}[t!]
\centering
\subfloat[][Inference time vs accuracy]{\includegraphics[width=0.3\textwidth]{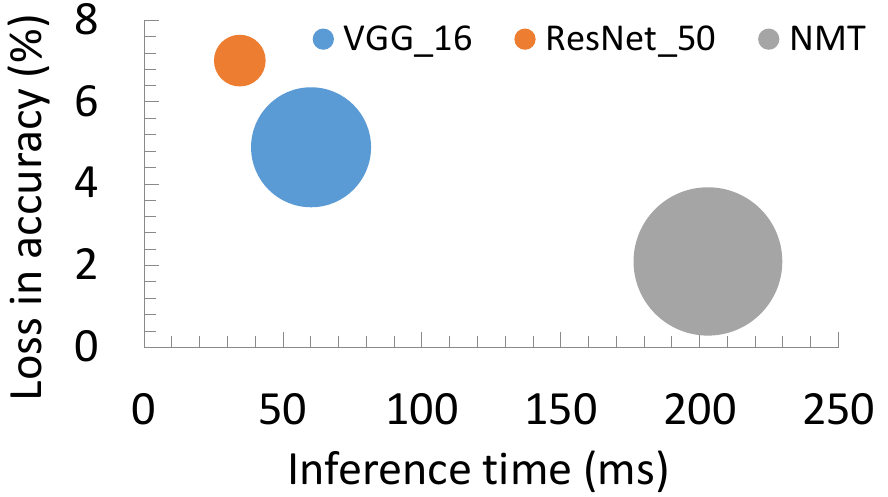}}
\hfill
\subfloat[][Energy vs accuracy]{\includegraphics[width=0.3\textwidth]{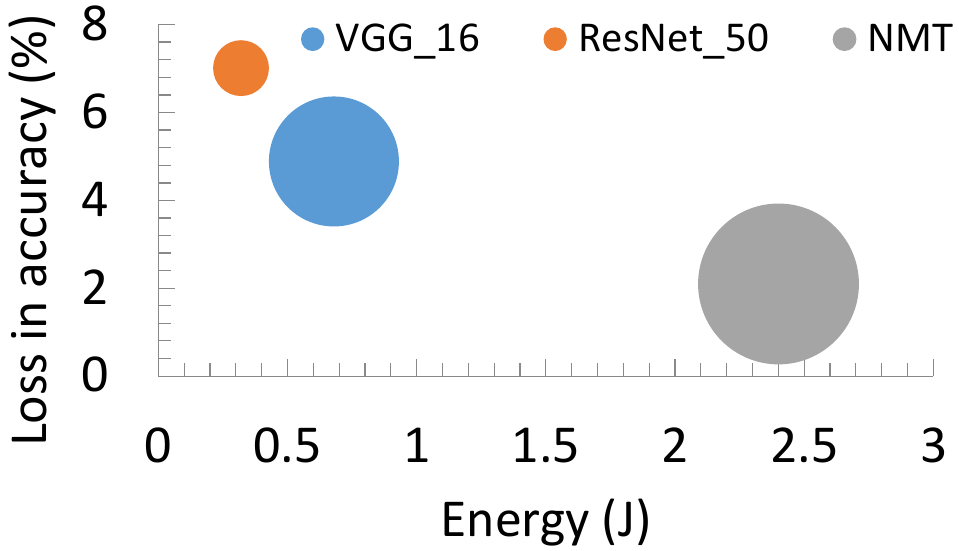}}
\hfill
\subfloat[][Compressed model size vs accuracy]{\includegraphics[width=0.3\textwidth]{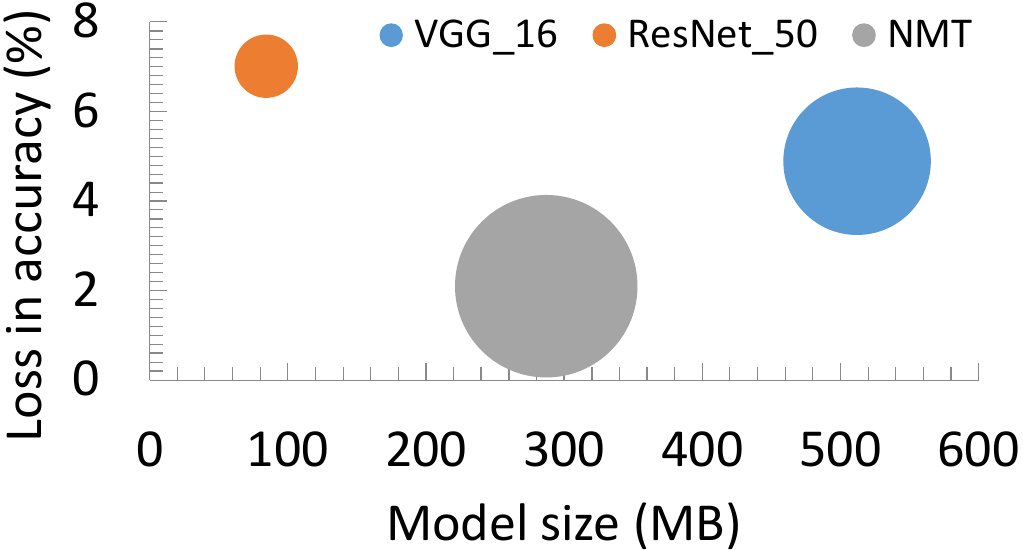}}
\hfill
\caption{ The \pruning effects on different sized models. The larger a bubble is, the more parameters the corresponding model has (see
Table~\ref{tab:workload}). } \label{fig:bubblepruning}
\vspace{-8mm}
\end{figure*}

\begin{figure*}[!t]
\centering
\subfloat[][Model size]{\includegraphics[width=0.24\textwidth]{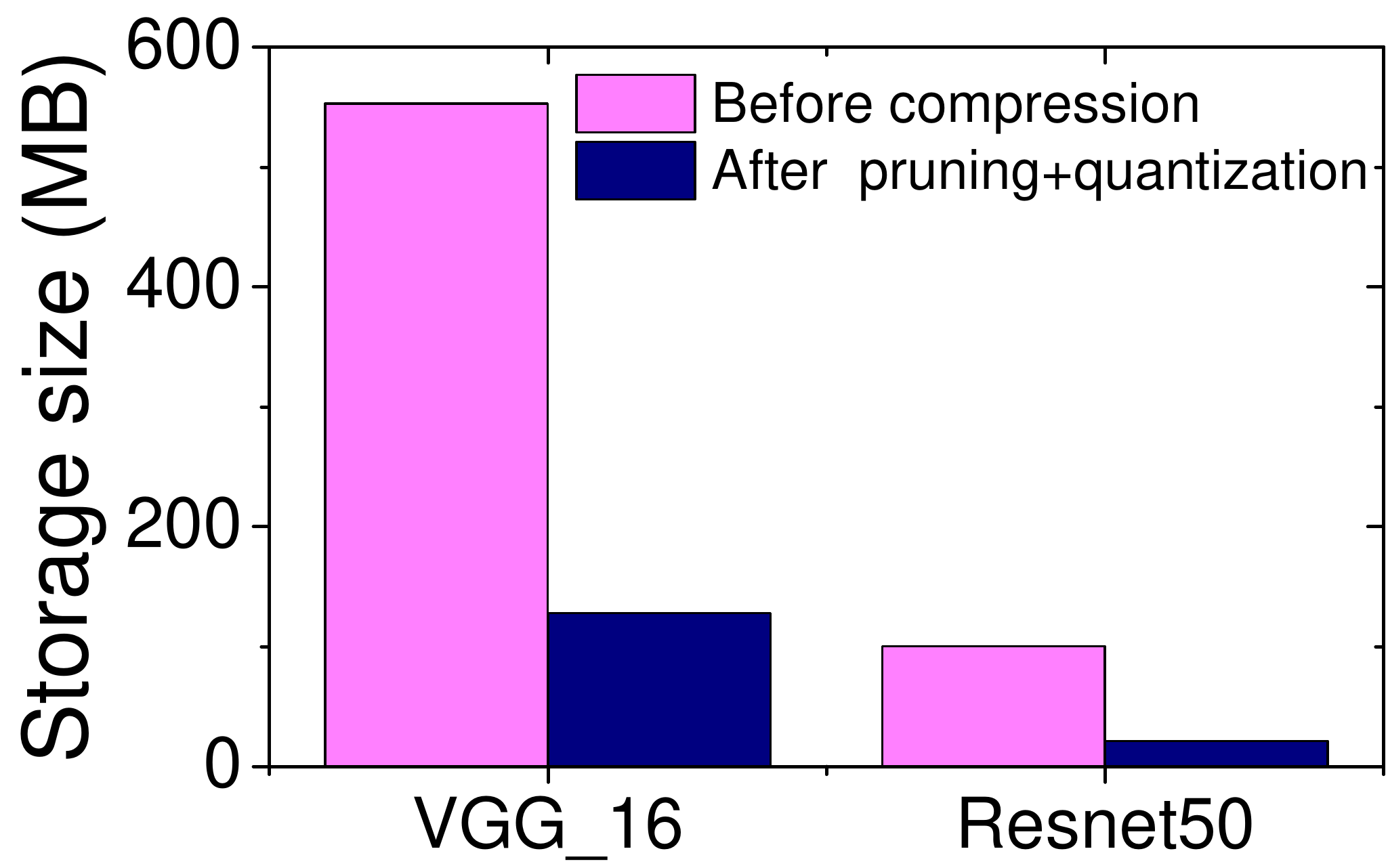}}
\hfill
\subfloat[][Inference time]{\includegraphics[width=0.24\textwidth]{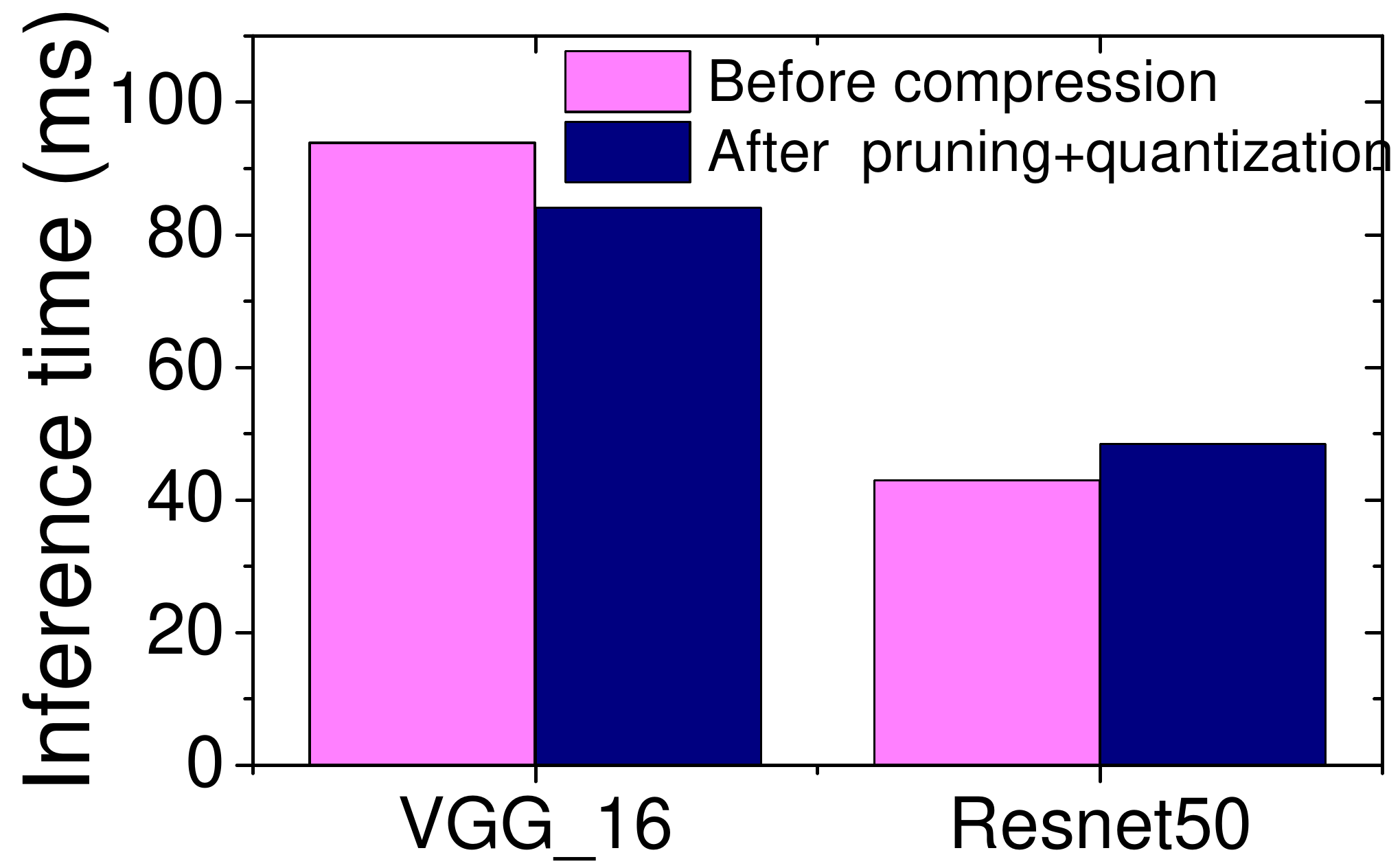}}
\hfill
\subfloat[][Energy consumption]{\includegraphics[width=0.24\textwidth]{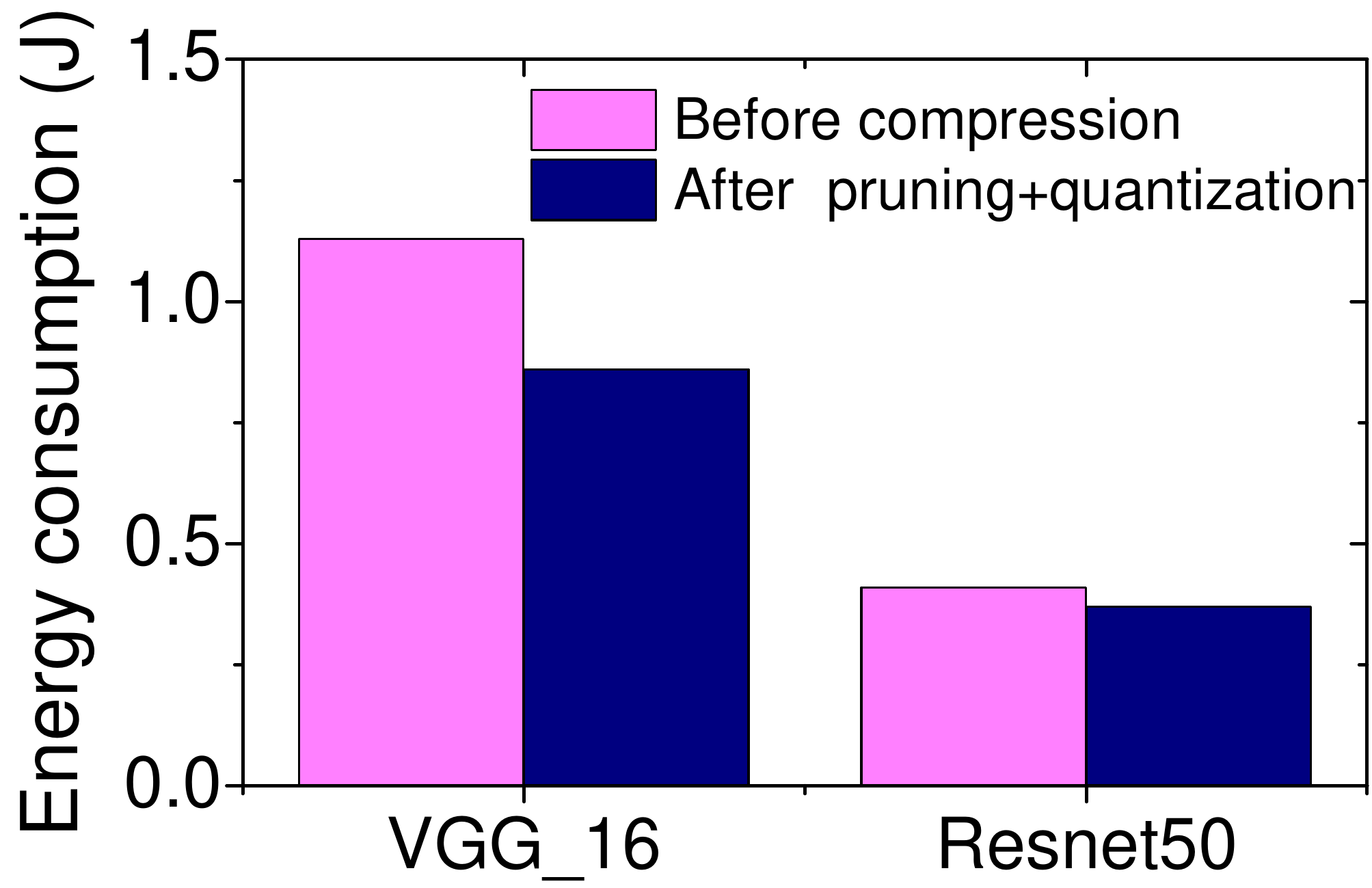}}
\hfill
\subfloat[][Accuracy]{\includegraphics[width=0.243\textwidth]{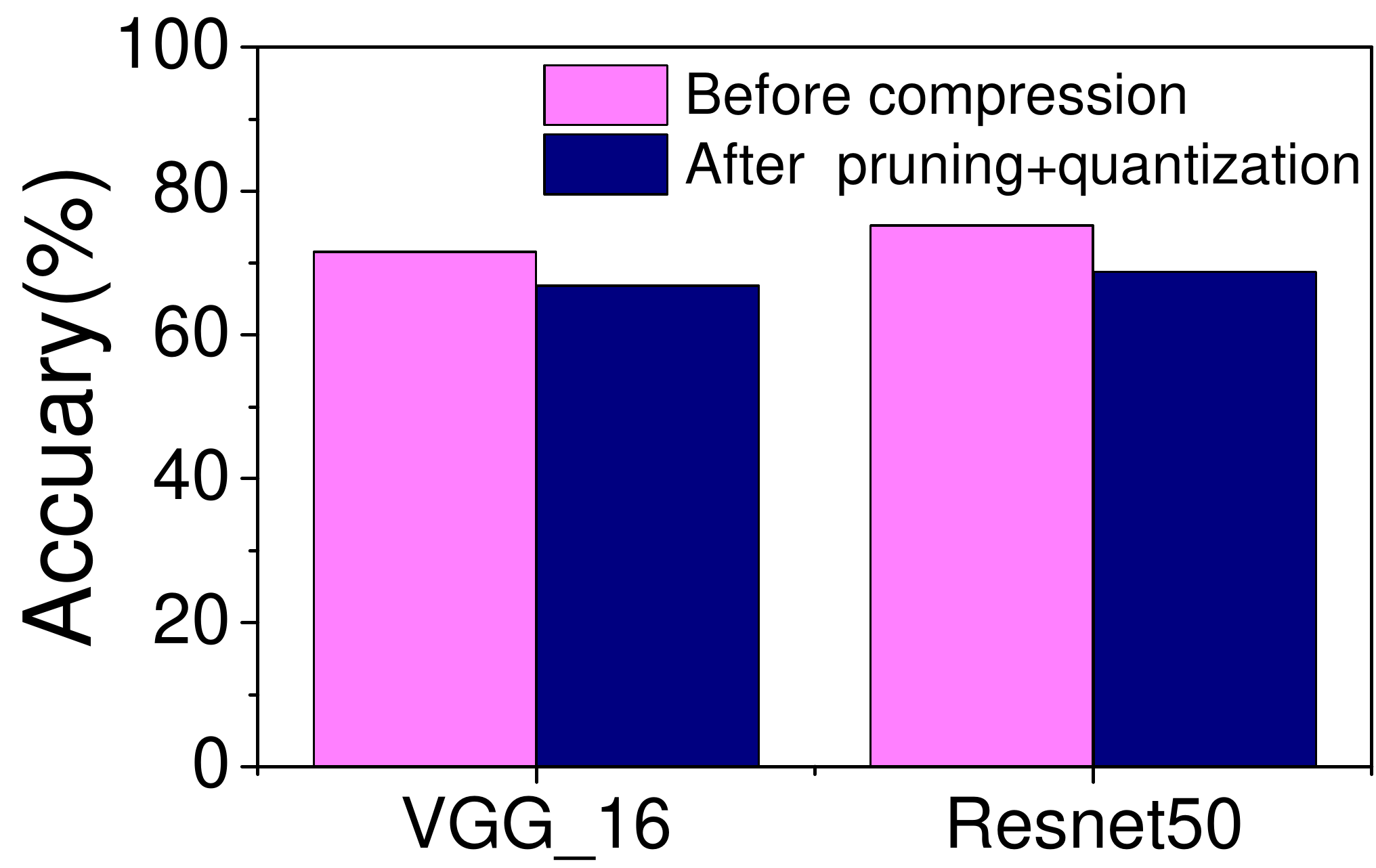}}
\hfill
\caption{The model size (a) inference time (b) energy consumption (c) and accuracy (d) before and after model compression.
}
\vspace{-5mm}
\label{fig:combine}
\end{figure*}

\subsection{Impact of Model Parameter Sizes}
The bubble charts in Figures~\ref{fig:bubblequantization} and \ref{fig:bubblepruning} quantify the impact by applying \quantization and
\pruning to the deep learning models of different sizes. Here, each bubble corresponds to a model. The size of the bubble is proportional
to the number of network parameters (see Table~\ref{tab:workload}). As can be seen from the diagrams, there is a non-trivial correlation
between the original model size, compression techniques, and the optimization constraints. This diagram suggests that there is a need for
adaptive schemes like parameter search to effectively explore the design space of multiple optimization objectives.





\subsection{Combining Pruning and Quantization\label{sec:combin}}

So far we have evaluated \pruning and \quantization in isolation. An natural question to ask is: ``Is it worthwhile to combine both
techniques?". Figure~\ref{fig:combine} shows the results by first applying a 8-bit \dquantization and then \pruning to \texttt{VGG\_16} and \texttt{ResNet50}.

As can be seen from Figure~\ref{fig:combine}a, combining both compression techniques can significantly reduce the model storage size -- the
resulting models are 76\% smaller than the original ones; and there is little degradation in the top-1 prediction accuracy
(Figure~\ref{fig:combine}d) -- less than 7\%. From Figure~\ref{fig:combine}b, we see that the combination has positive impact on the
inference time for \texttt{VGG\_16} as the runtime overhead of \dquantization (see Section~\ref{sec:time}) can be amortized by \pruning.
The combination, however, leads to longer inference time for \texttt{ResNet50} due to the expensive de-quantization overhead as we have
explained. Because of the difference in inference time, there is less benefit in energy consumpation for \texttt{ResNet50} over
\texttt{VGG\_16} (Figure~\ref{fig:combine}c). This experiment shows that combining \pruning and \quantization can be beneficial, but it
depends on the neural network architecture and what to optimize for.

%% file: discussions.tex
\section{Discussions}

Our evaluation reveals that \dquantization is particularly effective in reducing the model storage size and runtime memory footprint. As
such, it is attractive for devices with limited memory resources, particularly for small-formed IoT devices. However, \quantization leads
to longer inference time due to the overhead of the de-quantization process. Therefore, future research is needed to look at reducing the
overhead of de-quantization. We also observe that an 8-bit integer quantization seems to be a good trade-off between the model storage size
and the precision. This strategy also enables SIMD vectorization on CPUs and GPUs as multiple 8-bit scalar values can be packed into one
vector register. Using less than 8 bits is less beneficial on traditional CPUs and GPUs, but could be useful on FGPAs or specialized
hardware with purpose-built registers and memory load/store units. We believe studying when and how to apply \dquantization to a specific
domain or a neural network architecture would be an interesting research direction.

We empirically show that \pruning allows us to precisely control in the prediction precision. This is useful for applications like security
and surveillance where we need a degree of confidences on the predictive outcome. Compared to \dquantization, \pruning is less effective in
reducing the model storage size, and thus may require larger storage and memory space. We also find that \pruning is particularly effective
for \RNNs, perhaps due to the recurrent structures of an \RNN. This finding suggests \pruning can be an important method for accelerating
\RNN on embedded systems.

Combining \dquantization and \pruning is an interesting approach, as it can bring together the best part of both techniques (see
Section~\ref{sec:combin}). However, one must  make sure the overhead of \dquantization does not eclipse the reduction in inference time by
applying \pruning. One interesting research question could be: ``Can we find other data representations to better quantize a model?". For
examples, instead of using just integers, one can use a mixture of floating point numbers and integers with different bit widths – by
giving wider widths for more important weights. Furthermore, given that it is non-trivial to choose the right compression settings, it will
be very useful to have a tool to automatically search over the Pareto design space to find a good configuration to meet the conflict
requirements of the model size, inference time and prediction accuracy.  As a final remark of our discussion, we hope our work can
encourage a new line of research on auto-tuning of deep learning model compression.

%% file: related_work.tex
\section{Related Work}

There has been a significant amount of work on reducing the storage and computation work by model compression. These techniques include
pruning~\cite{Li2016Pruning}, quantization~\cite{Gong2014Compressing,han2015deep}, knowledge
distillation~\cite{hinton2015distilling,Sau2016Deep}, huffman coding~\cite{han2015deep}, low rank and sparse decomposition~\cite{denton2014exploiting}, decomposition~\cite{lebedev2014speeding}, etc. This paper develops a quantitative approach to
understand the cost and benefits of deep learning compression techniques. We target pruning and data quantization because these are widely
used and directly applicable to a pre-trained model.

In addition to model compression, other works exploit computation-offloading~\cite{teerapittayanon2017distributed,Kang2017neurosurgeon},
specialized hardware design~\cite{chen2017eyeriss,Han:2016:EEI:3001136.3001163}, and dynamic model
selection~\cite{Taylor:2018:ADL:3211332.3211336}. Our work aims to understand how to accelerate deep learning inference by choosing the
right model compression technique. Thus, these approaches are orthogonal to our work.

Off-loading computation to the cloud can accelerate \DNN model inference \cite{teerapittayanon2017distributed}. Neurosurgeon
\cite{Kang2017neurosurgeon} identifies when it is beneficial (\eg in terms of energy consumption and end-to-end latency) to offload a \DNN
layer to be computed on the cloud. The Pervasive \CNN~\cite{song2017towards} generates multiple computation kernels for each layer of a
\CNN, which are then dynamically selected according to the inputs and user constraints. A similar approach presented in
\cite{servia2017personal} trains a model twice, once on shared data and again on personal data, in an attempt to prevent personal data
being sent outside the personal domain. Computation off-loading is not always applicable due to privacy, latency or connectivity issues.
Our work is complementary to previous work on computation off-loading by offering insights to best optimize \emph{local} inference.


%% file: conclusions.tex
\section{Conclusions}
This paper has presented a comprehensive study to characterize the effectiveness of model compression techniques on embedded systems. We
consider two mainstream model compression techniques and apply them to a wide range of representative deep neural network architectures. We
show that there is no ``one-size-fits-all" universal compression setting, and the right decision depends on the target neural network
architecture and the optimization constraints. We reveal the cause of the performance disparity and demonstrate that a carefully chosen
parameter setting can lead to efficient embedded deep inference. We provide new insights and concrete guidelines, and define possible
avenues of research to enable efficient embedded inference.

%% file: ack.tex
\section*{Acknowledgments}
This work was supported in part by the NSF China under grant agreements 61872294 and 61602501; the Fundamental Research Funds for the
Central Universities under grant agreement GK201803063; the UK EPSRC through grant agreements EP/M01567X/1 (SANDeRs) and EP/M015793/1
(DIVIDEND); and the Royal Society International Collaboration Grant (IE161012). For any correspondence, please contact Jie Ren
(renjie@snnu.edu.cn), Jianbin Fang (j.fang@nudt.edu.cn) and Zheng Wang (z.wang@lancaster.ac.uk).